\documentclass{article}

\usepackage[table]{xcolor}
\usepackage[final]{corl_2026} 

\usepackage{multirow}
\usepackage{booktabs}
\usepackage{array}
\usepackage{amsmath}
\usepackage{amssymb}
\usepackage{algorithm}
\usepackage{algorithmic}
\usepackage{graphicx}
\usepackage{booktabs}
\usepackage{subcaption} 
\usepackage{adjustbox}
\usepackage{siunitx}
\usepackage{dsfont}
\usepackage{wrapfig}
\usepackage{enumitem}

\newcommand{\figref}[1]{Fig.~\ref{#1}}
\newcommand{\secref}[1]{Sec.~\ref{#1}}
\newcommand{\algref}[1]{Alg.~\ref{#1}}
\newcommand{\eqnref}[1]{Eq.~\eqref{#1}}
\newcommand{\tabref}[1]{Tab.~\ref{#1}}

\definecolor{metricWorse}{RGB}{248, 197, 195}
\definecolor{metricMid}{RGB}{255, 240, 195}
\definecolor{metricBetter}{RGB}{200, 232, 200}

\newcommand{\heat}[1]{%
  \cellcolor{%
    metricWorse!\fpeval{max(0,100*(1-2*(#1)))}%
    !metricMid!\fpeval{min(100,100*(2-2*(#1)))}%
    !metricBetter%
  }%
}

\newcommand{\metric}[2]{\heat{#1}#2}

\usepackage{tcolorbox}
\tcbset{
  kaminoapi/.style={
    colback=gray!4,
    colframe=gray!55,
    boxrule=0.4pt,
    arc=2pt,
    left=6pt, right=6pt, top=4pt, bottom=4pt,
    fontupper=\ttfamily\small,
  }
}

\title{Kamino: GPU-based Simulation of Rigid\\ Multi-Body Physics with Kinematic Loops}

\author{
  Vassilios Tsounis\textsuperscript{*\,1}\quad
  Guirec Maloisel\textsuperscript{*\,1}\quad
  Christian Schumacher\textsuperscript{*\,1}\quad
  Ruben Grandia\textsuperscript{1}\\[2pt]
  \bfseries
  Agon Serifi\textsuperscript{1}\quad
  David M{\"u}ller\textsuperscript{1}\quad
  Chris Amevor\textsuperscript{2}\quad
  Tobias Widmer\textsuperscript{2}\quad
  Moritz B{\"a}cher\textsuperscript{1}
  \AND
  \normalfont
  \textsuperscript{*}Joint first authors, \,\,
  \textsuperscript{1}Disney Research, Zurich, Switzerland, \,\,
  \textsuperscript{2}NVIDIA, Zurich, Switzerland\\[2pt]
  \texttt{vassilios.tsounis@disneyresearch.com}
}

\begin{document}
\maketitle


\begin{abstract}
We present Kamino\footnote{\url{https://kamino.disneyresearch.com/}}, a GPU-based physics solver for massively parallel simulations of heterogeneous highly-coupled mechanical systems. Implemented in Python using NVIDIA Warp and integrated into the Newton framework, it enables the application of data-driven methods, such as large-scale reinforcement learning, to complex robotic systems that exhibit strongly coupled kinematic and dynamic constraints such as kinematic loops. The latter are often circumvented by approximating the system's topology as a kinematic tree, and incorporating explicit loop-closure constraints or so-called mimic joints. Kamino aims at alleviating this burden by natively supporting these types of couplings.
Moreover, Kamino supports heterogeneous worlds, allowing for batched simulation of structurally diverse robots on a single GPU. 
At its core lies a state-of-the-art constrained optimization algorithm based on ADMM, which computes constraint forces by solving the constrained rigid multi-body forward dynamics transcribed as a nonlinear complementarity problem. This leads to high-fidelity simulations that resolve contact dynamics without resorting to approximate models that simplify or convexify the problem. We demonstrate RL policy training on a biped with six nested kinematic loops, generating a feasible walking policy while simulating $4{,}096$ parallel environments on a single GPU.
\end{abstract}


\begin{figure}[hb]
\centering
\includegraphics[width=1.0\linewidth]{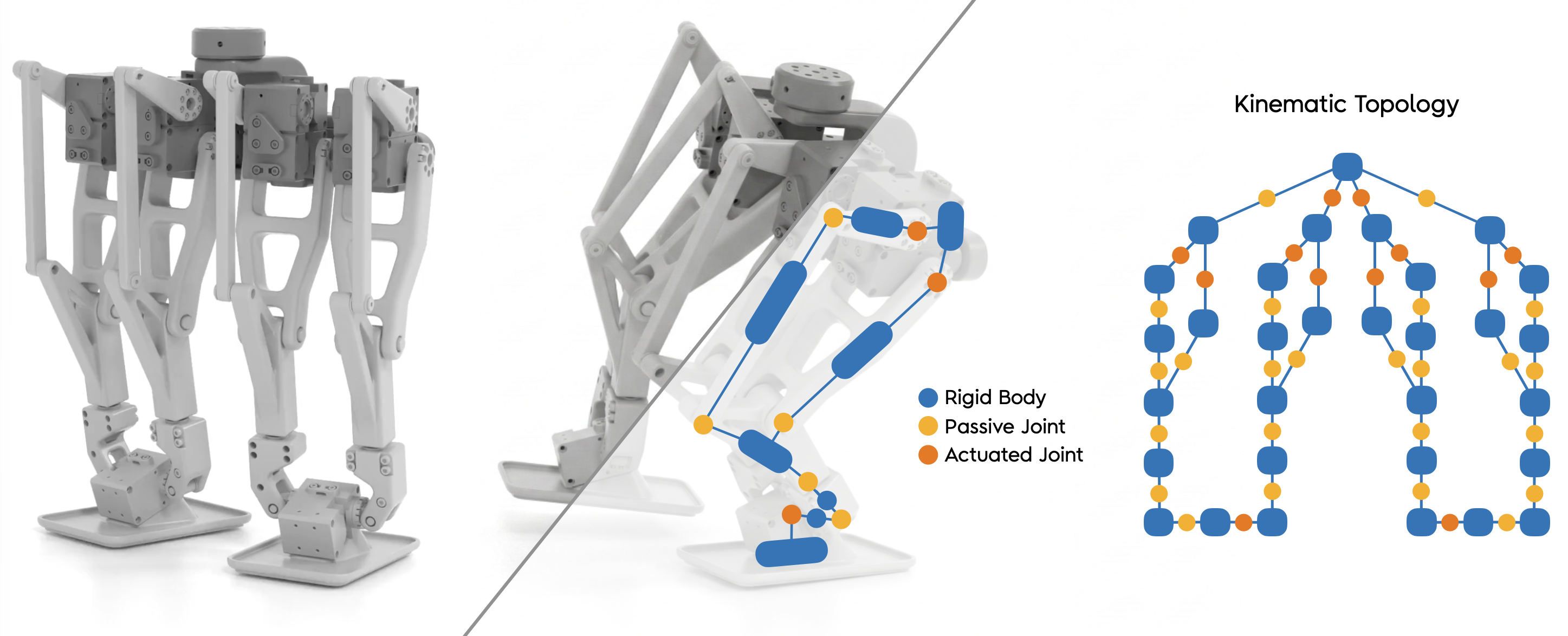}
\caption{DR~Legs (left) with kinematic graph overlaid for half of the left leg (middle). In the full kinematic graph (right), each leg contains two small kinematic loops nested in a third large one.}
\label{fig:overview}
\end{figure}

\section{Introduction}
\label{sec:introduction}

GPU-based physics simulation has revolutionized robot learning in recent years.
By running thousands of parallel environments entirely on-device, frameworks such as Isaac Gym~\cite{isaacgym2023}, Brax~\cite{freeman2021brax}, and MuJoCo~\cite{todorov2012mujoco}, including its GPU-accelerated variants MJX~\cite{mujoco_mjx_docs} and MuJoCo Warp (MJWarp)~\cite{mujoco_warp}, reduce Reinforcement Learning (RL) training times from days to minutes~\cite{rudin2022,mittal2023orbit}. 
The vast majority of these simulators rely on \emph{minimal-coordinate} formulations~\cite{featherstone2014rigid} that require a \emph{tree-structured} kinematic topology, leading to significantly reduced problem dimensions, and thus enabling the use of $O(n)$ recursive algorithms with purely local data dependencies --- an ideal case for GPU parallelism.

However, this tree-structured assumption excludes an important class of mechanisms: those with \textbf{kinematic loops}.
Closed kinematic loops arise whenever multiple paths in the kinematic graph connect the same pair of bodies, and are ubiquitous in parallel manipulators, legged robots with four-bar linkage transmissions~\cite{gim2018design}, and hydraulic excavators~\cite{egli2022general}.
While these designs offer superior power-to-weight ratios and structural rigidity, simulating them introduces severe inherent numerical challenges, such as algebraic ill-conditioning stemming from over-constrained topologies as well as couplings of bodies with large mass ratios.

Existing simulators typically handle loops by enforcing additional equality constraints over the kinematic tree. 
However, because closed loops are rarely the primary focus of these frameworks, comprehensive tooling for model specification, forward and inverse kinematics is often lacking. 
Furthermore, solving these systems requires tedious parameter tuning and injects inherent topological bias by arbitrarily splitting joints between the kinematic tree and equality constraints.

Beyond these topological complexities lies the critical requirement to accurately resolve the Nonlinear Complementarity Problem (NCP) that governs hard-contact rigid-body interactions.
Traditional GPU-accelerated simulators often resort to local, iterative methods like Projected Gauss--Seidel (PGS).
While computationally cheap per iteration, PGS struggles with the tight numerical coupling and high algebraic stiffness inherent to simultaneous closed loops and non-penetrating contacts, frequently resulting in constraint drift, artificial compliance, or outright solver divergence.
Other frameworks like MuJoCo~\cite{todorov2012mujoco} relax the non-smooth NCP into a strictly convex optimization problem, solved with quadratic convergence using a specialized Newton method. Similarly, Macklin et al.~\cite{Macklin_2019} introduced a non-smooth Newton method for GPU-based multi-body simulation that retains the NCP w.r.t. contacts, but still requires regularization to handle ill-conditioning.
However, such approaches rely on constraint compliance to maintain well-conditioned optimization gradients; as a result, simulating perfectly rigid joints and hard contacts can still induce numerical instabilities or require extensive parameter tuning.

Recent work has advanced both the modeling of mechanisms with kinematic loops and the solvers needed to accurately simulate them.
Schumacher et al.~\cite{schumacher2021versatile} developed inverse kinematics formulations for robots with kinematic loops. Brüdigam and Manchester~\cite{Br_digam_2021} introduced a linear-time variational integrator in maximal coordinates, which Howell et al.~\cite{howell2025dojodifferentiablephysicsengine} paired with an interior-point NCP solver in the differentiable physics engine Dojo. Maloisel et al.~\cite{maloisel2025versatile} proposed a quaternion-based constrained rigid-body dynamics framework, and Tsounis et al.~\cite{tsounis2025solving} provided a comprehensive comparison of forward dynamics solvers for constrained multi-body systems with frictional contacts.
In particular, \textit{proximal algorithms} such as the Alternating Direction Method of Multipliers (ADMM)~\cite{boyd2011admm} have emerged as highly accurate alternatives to PGS, demonstrating superior convergence properties and exceptional robustness when handling non-smooth contact dynamics and ill-conditioned problems. Tasora et al.~\cite{tasora2021admm} applied ADMM to variational inequalities in non-smooth dynamics, and Carpentier et al.~\cite{carpentier2021proximal, carpentier2024unified} developed proximal formulations that unify compliant and rigid contacts.
Yet, while numerical methods for high-fidelity closed-loop simulation are well-established, a GPU implementation capable of scaling these global optimization-based solvers for RL workloads has been missing.

Kamino fills this gap.
It is a GPU-native solver for constrained rigid multi-body systems with arbitrary joint topologies, built on a maximal-coordinate formulation~\cite{haug1989computer} where every body carries its own pose, and kinematic relationships are enforced as explicit algebraic constraints.
The forward dynamics problem is cast in the dual space of constraint reactions and solved via Proximal-ADMM (PADMM)~\cite{carpentier2024unified, tsounis2025solving}, naturally unifying bilateral joint constraints, unilateral joint limits, and frictional contacts into a single, robust optimization loop that accurately resolves the underlying NCP.
The global coupling introduced by kinematic loops in the Delassus matrix is handled by a linear solve, which constitutes the central computational challenge. It is addressed through either an efficient block-Cholesky decomposition (in the case of small systems) or a warm-started Conjugate Residual solver working with a block-sparse, matrix-free Delassus operator (in the case of large systems).
Moreover, due to its maximal-coordinate formulation, Kamino natively supports heterogeneous parallel worlds containing completely arbitrary and distinct robot morphologies. This capability opens new possibilities in batched simulation of structurally diverse robots and environments.

Kamino is integrated into the open-source Newton physics engine~\cite{nvidia2025newton}, built on NVIDIA Warp~\cite{warp2022}, which compiles Python-defined GPU kernels to CUDA and provides zero-copy tensor interoperability with PyTorch and JAX.
This integration allows Kamino to leverage Newton's core infrastructure and to be interfaced with directly in learning workflows via Isaac Lab~\cite{mittal2025isaac}, making its unique simulation capabilities available for RL.
We demonstrate RL policy training on DR~Legs~\cite{gim2018design}, a biped with complex topology, including multiple kinematic loops in each leg (see \figref{fig:overview}), using up to $4{,}096$~parallel environments on a single GPU.

In summary, and to the best of our knowledge, we contribute the first physics solver that simultaneously
\textbf{(i) handles arbitrary rigid multi-body systems with minimal parameter tuning}, including those exhibiting complex kinematic loops, by combining a maximal-coordinate formulation with a Proximal-ADMM dual solver;
\textbf{(ii) avoids approximations of the contact model} that introduce compliance or alter the physics;
\textbf{(iii) exploits massive parallelism on the GPU}, enabling efficient RL policy training;
\textbf{(iv) is available open-source}, as part of a flexible ecosystem that provides all necessary supporting functionality for learning workflows.
We evaluate our framework comprehensively, through rigorous validation of the physics and benchmarking, against state-of-the-art solvers across a broad spectrum of elementary mechanisms and complex robotic topologies. Kamino demonstrates the best overall physics fidelity, while scaling to thousands of parallel environments and integrating seamlessly with RL-based robotics frameworks.


\section{Method}
\label{sec:method}

This section provides a high-level overview of Kamino. We first outline its high-level API and integration into Newton (\secref{sec:api}). We then introduce the core formulation governing constrained rigid multi-body dynamics in maximal coordinates (\secref{sec:dynamics}). Finally, we describe the Proximal-ADMM solver used to efficiently resolve these tightly coupled constraints on the GPU (\secref{sec:solver}). Expanded implementation details, optimizations, and algorithmic pseudocode are provided in the Appendix, while an exhaustive mathematical treatment of the constrained multi-body formulation can be found in the extended technical report~\cite{tsounis2025solving}.

\subsection{High-level API}
\label{sec:api}
Kamino is exposed in Newton as a configurable physics backend through \texttt{SolverKamino}, which implements the common \texttt{SolverBase} interface shared by all Newton solvers. 
The simulation is described by a static \texttt{Model} (geometry, inertia, joints, actuators) and time-varying \texttt{State}, \texttt{Control} and \texttt{Contacts} data containers allocated on the GPU. At each time step, the host calls
\begin{tcolorbox}[kaminoapi, title={Simulation step (per time step $\Delta t$)}]
solver.step(state\_in, state\_out, control, contacts, dt)
\end{tcolorbox}
which advances the dynamics by $\Delta t$: it reads the current configuration and controls, solves the constrained forward-dynamics problem, integrates the result, and writes the next state to \texttt{state\_out}. 

A standard double-buffering pattern swaps \texttt{state\_in} and \texttt{state\_out} after each step, enabling in-place GPU pipelines and CUDA graph capture. Contacts may be supplied from Newton's shared collision pipeline or, optionally, generated by Kamino's internal detector.

In addition to forward simulation, RL applications require the ability to reset the robot to a (randomized) state. For articulated systems, resets are straightforward due to the minimal state representation: a user can directly modify the generalized coordinates and velocities. However, for maximal-coordinate representations, especially for systems with closed loops, manually constructing a kinematically consistent state is challenging. Kamino, therefore, additionally provides
\begin{tcolorbox}[kaminoapi, title={Simulation reset}]
solver.reset(state, world\_mask, \ldots)
\end{tcolorbox}
which writes a consistent initial configuration into \texttt{state}. Resets can be specified through joint coordinates, actuated DoFs, base body states, or full body states, with an optional per-world mask to support asynchronous multi-environment applications. When only joint-level states are specified, a dedicated forward-kinematics solver reconciles redundant or loop-closure mechanisms so that body poses satisfy all kinematic constraints~\cite{schumacher2021versatile}. 
Finally, solver-specific parameters (integrator, constraint stabilization, tolerances) are grouped in \texttt{SolverKamino.Config}.
Together, this API lets Kamino reuse Newton's asset pipeline, collision module, and visualization while substituting only the dynamics and constraint solution at \texttt{step} time.


\subsection{Constrained Dynamics in Maximal Coordinates}
\label{sec:dynamics}

Kamino adopts a maximal-coordinate formulation~\cite{haug1989computer} in which each rigid body $i$ carries an independent SE(3) pose $\mathbf{q}_i$ and a 6D spatial twist $\mathbf{u}_i$. The equations of motion follow the \textit{impulse--velocity} Newton--Euler formulation: $\mathbf{M}\left(\mathbf{u}^{+} - \mathbf{u}^{-}\right) = \Delta{t}\,\mathbf{h} + \mathbf{J}^\top\boldsymbol{\lambda}$, where $\mathbf{M}(\mathbf{q})$ is the block-diagonal mass matrix, $\mathbf{h}(\mathbf{q},\mathbf{u})$ comprises gravitational, gyroscopic and external forces, $\mathbf{J}(\mathbf{q})$ is the constraint Jacobian matrix, and $\boldsymbol{\lambda}$ defines the constraint reactions vector.
%

Three categories of constraints are unified into a single system: (i) \emph{Bilateral joint constraints} 
enforce kinematic relationships between pairs of bodies, or a body and the world~\cite{maloisel2025versatile}, each locking up to $6$ degrees of freedom depending on the joint type.
%
%
For actuated joints, implicit PD control is supported, where proportional-derivative gains are absorbed into the constraint system for additional numerical stability. Joint reflected inertia, often referred to as \textit{armature}, as well as viscous joint damping, are handled similarly.
(ii) \emph{Unilateral joint limits} enforce inequality bounds on joint coordinates. 
(iii) \emph{Contacts} are modeled via Signorini-Coulomb-Newton complementarity with the de~Saxcé correction~\cite{desaxce1998bipotential, acary2011formulation}, which reformulates the non-associated Coulomb friction law as a complementarity condition on the Cartesian product of Coulomb cones. 
The detailed constraint formulation, including stabilization via Baumgarte bias terms~\cite{baumgarte1972stabilization}, follows~\cite{tsounis2025solving}.

At each time step, the equations of motion are advanced using a discrete-time integration scheme. Gathering the resulting system and eliminating the next-step velocity variables via a Schur complement reduces the forward dynamics to a \emph{dual problem in constraint reactions}~\cite{tsounis2025solving}:
\begin{equation}
  \boldsymbol{\lambda} = \operatorname*{argmin}_{\mathbf{x} \in \mathcal{K}} \;
  \tfrac{1}{2}\, \mathbf{x}^\top \mathbf{D}\, \mathbf{x}
  + \mathbf{x}^\top \bigl(\mathbf{v}_f + \boldsymbol{\Gamma}(\mathbf{v}^+(\mathbf{x}))\bigr),
  \label{eq:dual_problem}
\end{equation}
where $\boldsymbol{\Gamma}(\cdot)$ is the de~Saxcé correction operator, 
$\mathbf{D}$
is the Delassus matrix, and $\mathbf{v}_f$ is the free velocity vector, both detailed in Appendix~\ref{app:dynamics}.
The solution is restricted to the feasible cone~$\mathcal{K}$.
The de~Saxcé term renders the objective nonlinear and is a crucial part of formulating the optimization problem that describes rigid-body and hard-contact models. The resulting velocity offset eliminates non-zero velocities along the contact normals, ensuring that sliding contacts do not push bodies into or away from each other.

Existing GPU simulators typically require homogeneous environments where all worlds share the same number of simulation entities, such as bodies and joints.
Kamino lifts this restriction.
All GPU kernels operate on 2D thread grids of shape $(\texttt{num\_worlds}, \texttt{max\_dim})$, where $\texttt{max\_dim}$ is the maximum over all worlds of the relevant quantity (bodies, constraints, contacts, etc.).
Each world stores its actual dimension, and threads whose index exceeds the per-world count exit early via a mask check.


\subsection{Proximal-ADMM Solver}
\label{sec:solver}

The dual problem~\eqnref{eq:dual_problem} is a non-smooth, cone-constrained optimization problem.
Following~\cite{tsounis2025solving, boyd2011admm, carpentier2021proximal}, we solve it via Proximal-ADMM (PADMM), which decouples the global physical coupling of the Delassus matrix from the non-smooth contact formulation via an alternating sequence of primal, consensus, and dual updates.
Within this iterative loop, the most significant computational bottleneck lies in the inner linear system solve performed at each iteration to handle the global constraint interactions.
Thus, its implementation directly determines how the solver throughput and memory footprint scale with the problem size and number of parallel worlds.   

Kamino provides a few options to the user that control both of these aspects. As the first option, the constraint Jacobian $\mathbf{J}$ can be represented as either a dense or sparse matrix. Second, two approaches to the linear system solve are offered: a) a \textit{dense} formulation that employs block-Cholesky factorization (LLT) of $\mathbf{D}$, and b) a \textit{sparse} formulation that realizes a matrix-free Delassus matrix-vector operation used in conjunction with a Conjugate Residual (CR) iterative solver~\cite{Macklin_2019}. Which representation is the most appropriate to use depends on the overall system size as well as the expected density of contact constraints compared to those of joints.

Additionally, several other techniques are combined to make the iterative solver perform well in practice for parallel simulation on the GPU. These and other details about the solver can be found in Appendix~\ref{app:solver}, including the solver pseudocode in \algref{alg:padmm}.


\section{Experimental Results}
\label{sec:results}

\begin{figure}[tb]
    \centering
    \includegraphics[width=\linewidth]{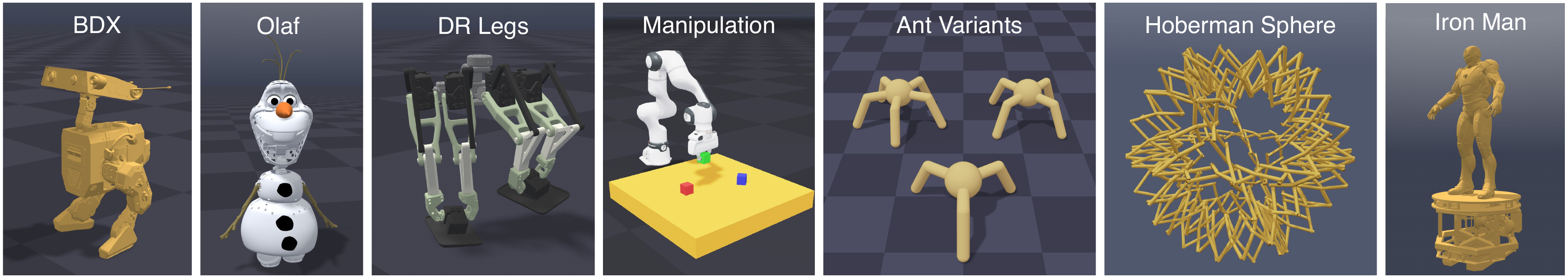}
    \caption{\textbf{Simulation Models} used in our experiments are (from left to right): BDX Droid, Olaf robot, DR Legs, two-finger robotic arm, Ant variants (3-, 4-, and 5-legged morphologies), Hoberman Sphere, and the Iron Man Audio-Animatronics\textregistered\ figure. \textcopyright\ MARVEL, Disneyland Paris.}
    \label{fig:systems}
\end{figure}

We evaluate Kamino (\texttt{SolverKamino}) against other state-of-the-art GPU-based physics solvers\footnote{All GPU-based Newton solvers used in this work only support $32$-bit single-precision floating-point data and operations, while MuJoCo on CPU is used with its default setting of $64$-bit double-precision.} available in the Newton framework, namely MJWarp (\texttt{SolverMuJoCo}) and XPBD (\texttt{SolverXPBD}), as well as MuJoCo on CPU. 
We assess performance across a diverse set of mechanical and robotic systems, beginning with elementary examples to validate simulator fidelity and extending to more complex robotic systems, including DR~Legs~\cite{gim2018design}, BDX Droid~\cite{ruben2024design}, Olaf robot~\cite{muller2026olaf}, and the Iron Man Audio-Animatronics\textregistered\ figure. These simulation models span both tree-structured and closed-loop kinematic topologies. We additionally evaluate variants of the classic Ant~\cite{Brockman2016} model, with different morphologies across parallel worlds to demonstrate Kamino's ability to simulate heterogeneous environments. The different simulation models are shown in~\figref{fig:systems} and a detailed overview is provided in~\secref{app:sim_models_overview}.

\subsection{Physics Validation}
\label{sec:results:validation}

\paragraph{Elementary Examples}
We validate the core building blocks of our formulation on four elementary examples, evaluating how accurately individual physical phenomena are captured. Specifically, we examine maximally dissipative hard frictional contact using a sliding box, Newtonian impacts using a bouncing ball, robustness to large mass ratios using stacked boxes, and closed-loop kinematics using a passive four-bar linkage (see~\figref{fig:validation_examples}). A key advantage of these targeted experiments is that they admit analytical reference solutions derived from first principles.
Details and numerical results are provided in Appendix~\ref{app:elementary_examples}.

Across all scenarios, Kamino closely matches the analytical solution without requiring example-specific tuning. The only tuned parameter is the Baumgarte stabilization weight in the kinematic-loop experiment to balance constraint satisfaction and energy preservation. In contrast, the other solvers exhibit distinct physical and numerical trade-offs. 
MJWarp introduces nonlinear softening into constraints, tightly coupling physical behavior and numerical stability, and, therefore, requiring careful parameter tuning. As a result, it cannot simultaneously reproduce physically correct stick-slip transitions and restitution without significant ground penetration. It also exhibits contact penetration under heavy loads, and the absence of the de~Saxcé correction in the contact-force formulation leads to artificial wobbling during sliding. In closed-loop systems, the underlying spring dynamics can introduce resonances that inject energy into the system.
XPBD, on the other hand, makes fewer modeling approximations but relies on localized PGS-like iterations rather than a global linear solve. Consequently, it requires a large number of iterations to recover physically accurate behavior, particularly in the presence of large mass ratios. Moreover, its position-level formulation cannot realize true sticking contacts, resulting in premature sliding, and introduces excessive energy dissipation at joint constraints. Overall, these examples demonstrate that Kamino’s PADMM-based formulation successfully avoids these limitations, capturing high-fidelity physics without relying on compliant constraints or high iteration counts.

\paragraph{Manipulation}

We evaluate Kamino on a cube-stacking manipulation task using a two-finger robotic arm (\figref{fig:systems}). The robot uses an inverse-kinematics controller driving PD position setpoints alongside feedforward gravity compensation. The system achieved a \SI{100}{\percent} success rate across $100$ runs with randomized initial conditions, demonstrating that Kamino produces the contact stability required for manipulation.

\paragraph{Physical Accuracy on Robotic Systems}
\label{sec:results:metrics}

\begin{table}[t]
\centering
\caption{\textbf{Physical accuracy comparison.} 
Performance metrics (see Appendix~\ref{app:metrics}) evaluated across models and solvers. We use \emph{median} and \emph{90th-percentile} aggregation over time, to filter out challenging frames which present ground penetration upon fast impacts (no continuous collision detection), causing large residuals for all solvers. Collisions are not modeled for the Iron Man example. Configuration-level metrics ($r_{J}$, $r_{C}$) are evaluated with each solver running independent trajectories, to capture error accumulation over time. Other metrics, which measure the single-step accuracy, are evaluated on single-step problems, the \emph{same} for each solver, sampled along a reference trajectory obtained by running Kamino at 10 kHz. We verified that the configuration-level residuals along the reference trajectory are low enough (largest p90 of $3{\cdot}10^{-4}$ across all examples) to provide valid starting points for all solvers. We use $20$ XPBD iterations for most examples, and $200$ for Iron Man, as the solver diverges with default settings. We use 200 PADMM iterations for Kamino. \\\hspace{\textwidth}\\\hspace{\textwidth}* We do not measure $r_{primal}, r_{cmpl}, r_{vi}$ for MuJoCo CPU as contact forces are unavailable.}
\label{tab:perf-by-system-stats}
\scriptsize
\renewcommand{\arraystretch}{1.15}
\setlength{\tabcolsep}{2pt}
\begin{tabular}{@{}l l c c c c c c c c@{}}
	\toprule
		& & \multicolumn{2}{c}{\textbf{Kamino}} & \multicolumn{2}{c}{\textbf{MJWarp}} & \multicolumn{2}{c}{\textbf{MuJoCo CPU$^*$}} & \multicolumn{2}{c}{\textbf{XPBD}} \\
		\cmidrule(lr){3-4}\cmidrule(lr){5-6}\cmidrule(lr){7-8}\cmidrule(lr){9-10}
		Model & Metric & median & p90 & median & p90 & median & p90 & median & p90 \\
		\midrule
		\multirow{8}{*}{\textbf{BDX}}
		& $r_{J}$
		& \metric{1.0}{$1.12{\cdot}10^{-4}$} & \metric{1.0}{$5.72{\cdot}10^{-4}$} & \metric{1.0}{$1.63{\cdot}10^{-7}$} & \metric{1.0}{$2.19{\cdot}10^{-7}$} & \metric{1.0}{$1.48{\cdot}10^{-7}$} & \metric{1.0}{$2.00{\cdot}10^{-7}$} & \metric{0.54}{$6.12{\cdot}10^{-3}$} & \metric{1.0}{$9.99{\cdot}10^{-3}$} \\
		& $r_{kin}$
		& \metric{1.0}{$5.90{\cdot}10^{-4}$} & \metric{1.0}{$1.63{\cdot}10^{-3}$} & \metric{0.42}{$9.60{\cdot}10^{-2}$} & \metric{0.73}{$2.92{\cdot}10^{-1}$} & \metric{0.42}{$9.82{\cdot}10^{-2}$} & \metric{0.72}{$2.95{\cdot}10^{-1}$} & \metric{0.0}{$6.51{\cdot}10^{0}$} & \metric{0.0}{$9.86{\cdot}10^{0}$} \\
		& $r_{C}$
		& \metric{0.72}{$2.95{\cdot}10^{-4}$} & \metric{1.0}{$3.62{\cdot}10^{-4}$} & \metric{0.0}{$3.05{\cdot}10^{-2}$} & \metric{0.1}{$3.43{\cdot}10^{-2}$} & \metric{0.17}{$2.55{\cdot}10^{-3}$} & \metric{0.48}{$7.75{\cdot}10^{-3}$} & \metric{0.81}{$2.06{\cdot}10^{-4}$} & \metric{1.0}{$2.22{\cdot}10^{-4}$} \\
		& $r_{v}$
		& \metric{1.0}{$3.05{\cdot}10^{-7}$} & \metric{0.57}{$1.84{\cdot}10^{-1}$} & \metric{0.76}{$1.02{\cdot}10^{-2}$} & \metric{0.52}{$2.10{\cdot}10^{-1}$} & \metric{0.63}{$1.51{\cdot}10^{-2}$} & \metric{0.54}{$2.01{\cdot}10^{-1}$} & \metric{0.8}{$9.06{\cdot}10^{-3}$} & \metric{0.54}{$2.00{\cdot}10^{-1}$} \\
		& $r_{primal}$
		& \metric{1.0}{$0.00{\cdot}10^{0}$} & \metric{1.0}{$3.73{\cdot}10^{-9}$} & \metric{1.0}{$0.00{\cdot}10^{0}$} & \metric{1.0}{$1.49{\cdot}10^{-8}$} & - & - & \metric{1.0}{$0.00{\cdot}10^{0}$} & \metric{1.0}{$0.00{\cdot}10^{0}$} \\
		& $r_{dual}$
		& \metric{1.0}{$2.09{\cdot}10^{-7}$} & \metric{0.7}{$1.23{\cdot}10^{-1}$} & \metric{0.9}{$6.84{\cdot}10^{-3}$} & \metric{0.65}{$1.43{\cdot}10^{-1}$} & \metric{0.76}{$1.01{\cdot}10^{-2}$} & \metric{0.67}{$1.35{\cdot}10^{-1}$} & \metric{0.93}{$6.08{\cdot}10^{-3}$} & \metric{0.67}{$1.35{\cdot}10^{-1}$} \\
		& $r_{cmpl}$
		& \metric{1.0}{$8.35{\cdot}10^{-8}$} & \metric{1.0}{$1.29{\cdot}10^{-5}$} & \metric{0.0}{$1.24{\cdot}10^{-2}$} & \metric{0.0}{$1.34{\cdot}10^{-1}$} & - & - & \metric{0.15}{$2.83{\cdot}10^{-3}$} & \metric{0.18}{$2.44{\cdot}10^{-2}$} \\
		& $r_{vi}$
		& \metric{1.0}{$3.03{\cdot}10^{-4}$} & \metric{0.7}{$1.23{\cdot}10^{-1}$} & \metric{0.11}{$7.23{\cdot}10^{-2}$} & \metric{0.46}{$2.56{\cdot}10^{-1}$} & - & - & \metric{0.51}{$2.20{\cdot}10^{-2}$} & \metric{0.58}{$1.74{\cdot}10^{-1}$} \\
		\midrule
		\multirow{8}{*}{\textbf{Olaf}}
		& $r_{J}$
		& \metric{1.0}{$1.40{\cdot}10^{-4}$} & \metric{1.0}{$8.30{\cdot}10^{-4}$} & \metric{1.0}{$9.54{\cdot}10^{-4}$} & \metric{1.0}{$2.02{\cdot}10^{-3}$} & \metric{0.99}{$1.02{\cdot}10^{-3}$} & \metric{1.0}{$2.82{\cdot}10^{-3}$} & \metric{0.0}{$5.38{\cdot}10^{-2}$} & \metric{0.33}{$1.38{\cdot}10^{-1}$} \\
		& $r_{kin}$
		& \metric{1.0}{$4.66{\cdot}10^{-3}$} & \metric{1.0}{$6.51{\cdot}10^{-2}$} & \metric{1.0}{$9.42{\cdot}10^{-3}$} & \metric{1.0}{$5.29{\cdot}10^{-2}$} & \metric{1.0}{$9.46{\cdot}10^{-3}$} & \metric{1.0}{$5.23{\cdot}10^{-2}$} & \metric{0.0}{$1.10{\cdot}10^{0}$} & \metric{0.3}{$1.56{\cdot}10^{0}$} \\
		& $r_{C}$
		& \metric{0.72}{$2.97{\cdot}10^{-4}$} & \metric{1.0}{$3.05{\cdot}10^{-4}$} & \metric{0.0}{$8.08{\cdot}10^{-2}$} & \metric{0.0}{$8.09{\cdot}10^{-2}$} & \metric{0.0}{$4.91{\cdot}10^{-2}$} & \metric{0.0}{$9.70{\cdot}10^{-2}$} & \metric{0.82}{$2.06{\cdot}10^{-4}$} & \metric{1.0}{$2.81{\cdot}10^{-4}$} \\
		& $r_{v}$
		& \metric{1.0}{$9.19{\cdot}10^{-7}$} & \metric{1.0}{$8.39{\cdot}10^{-6}$} & \metric{1.0}{$0.00{\cdot}10^{0}$} & \metric{0.97}{$5.40{\cdot}10^{-2}$} & \metric{0.79}{$9.50{\cdot}10^{-3}$} & \metric{0.9}{$6.83{\cdot}10^{-2}$} & \metric{0.88}{$7.20{\cdot}10^{-3}$} & \metric{1.0}{$1.31{\cdot}10^{-2}$} \\
		& $r_{primal}$
		& \metric{1.0}{$0.00{\cdot}10^{0}$} & \metric{1.0}{$4.66{\cdot}10^{-10}$} & \metric{1.0}{$0.00{\cdot}10^{0}$} & \metric{1.0}{$1.86{\cdot}10^{-9}$} & - & - & \metric{1.0}{$0.00{\cdot}10^{0}$} & \metric{1.0}{$1.46{\cdot}10^{-11}$} \\
		& $r_{dual}$
		& \metric{1.0}{$6.15{\cdot}10^{-7}$} & \metric{1.0}{$5.63{\cdot}10^{-6}$} & \metric{1.0}{$0.00{\cdot}10^{0}$} & \metric{1.0}{$3.92{\cdot}10^{-2}$} & \metric{0.89}{$7.00{\cdot}10^{-3}$} & \metric{0.98}{$5.24{\cdot}10^{-2}$} & \metric{1.0}{$4.83{\cdot}10^{-3}$} & \metric{1.0}{$8.81{\cdot}10^{-3}$} \\
		& $r_{cmpl}$
		& \metric{1.0}{$3.73{\cdot}10^{-6}$} & \metric{1.0}{$1.95{\cdot}10^{-5}$} & \metric{0.17}{$2.54{\cdot}10^{-3}$} & \metric{0.25}{$1.90{\cdot}10^{-2}$} & - & - & \metric{0.0}{$8.38{\cdot}10^{-3}$} & \metric{0.18}{$2.48{\cdot}10^{-2}$} \\
		& $r_{vi}$
		& \metric{1.0}{$7.58{\cdot}10^{-5}$} & \metric{1.0}{$4.26{\cdot}10^{-4}$} & \metric{0.3}{$4.06{\cdot}10^{-2}$} & \metric{0.8}{$9.11{\cdot}10^{-2}$} & - & - & \metric{0.43}{$2.77{\cdot}10^{-2}$} & \metric{1.0}{$3.29{\cdot}10^{-2}$} \\
		\midrule
		\multirow{8}{*}{\textbf{DR Legs}}
		& $r_{J}$
		& \metric{1.0}{$1.89{\cdot}10^{-4}$} & \metric{1.0}{$1.41{\cdot}10^{-3}$} & \metric{0.17}{$2.52{\cdot}10^{-2}$} & \metric{0.7}{$3.20{\cdot}10^{-2}$} & \metric{0.22}{$2.15{\cdot}10^{-2}$} & \metric{0.72}{$3.03{\cdot}10^{-2}$} & \metric{0.6}{$4.71{\cdot}10^{-3}$} & \metric{1.0}{$7.33{\cdot}10^{-3}$} \\
		& $r_{kin}$
		& \metric{1.0}{$1.05{\cdot}10^{-4}$} & \metric{1.0}{$4.01{\cdot}10^{-4}$} & \metric{0.69}{$3.41{\cdot}10^{-2}$} & \metric{0.93}{$1.32{\cdot}10^{-1}$} & \metric{0.92}{$1.37{\cdot}10^{-2}$} & \metric{1.0}{$8.23{\cdot}10^{-2}$} & \metric{0.0}{$1.84{\cdot}10^{0}$} & \metric{0.05}{$4.18{\cdot}10^{0}$} \\
		& $r_{C}$
		& \metric{1.0}{$8.97{\cdot}10^{-5}$} & \metric{1.0}{$1.25{\cdot}10^{-4}$} & \metric{0.48}{$7.56{\cdot}10^{-4}$} & \metric{0.74}{$2.73{\cdot}10^{-3}$} & \metric{0.48}{$7.52{\cdot}10^{-4}$} & \metric{0.85}{$1.78{\cdot}10^{-3}$} & \metric{1.0}{$2.43{\cdot}10^{-5}$} & \metric{1.0}{$3.11{\cdot}10^{-5}$} \\
		& $r_{v}$
		& \metric{1.0}{$6.92{\cdot}10^{-4}$} & \metric{0.08}{$7.92{\cdot}10^{-1}$} & \metric{0.9}{$6.73{\cdot}10^{-3}$} & \metric{0.08}{$7.90{\cdot}10^{-1}$} & \metric{0.65}{$1.45{\cdot}10^{-2}$} & \metric{0.08}{$7.90{\cdot}10^{-1}$} & \metric{0.61}{$1.62{\cdot}10^{-2}$} & \metric{0.16}{$6.16{\cdot}10^{-1}$} \\
		& $r_{primal}$
		& \metric{1.0}{$0.00{\cdot}10^{0}$} & \metric{1.0}{$4.66{\cdot}10^{-10}$} & \metric{1.0}{$0.00{\cdot}10^{0}$} & \metric{1.0}{$0.00{\cdot}10^{0}$} & - & - & \metric{1.0}{$0.00{\cdot}10^{0}$} & \metric{1.0}{$2.91{\cdot}10^{-11}$} \\
		& $r_{dual}$
		& \metric{1.0}{$6.83{\cdot}10^{-4}$} & \metric{0.27}{$4.43{\cdot}10^{-1}$} & \metric{0.91}{$6.52{\cdot}10^{-3}$} & \metric{0.27}{$4.40{\cdot}10^{-1}$} & \metric{0.66}{$1.40{\cdot}10^{-2}$} & \metric{0.27}{$4.40{\cdot}10^{-1}$} & \metric{0.82}{$8.55{\cdot}10^{-3}$} & \metric{0.36}{$3.38{\cdot}10^{-1}$} \\
		& $r_{cmpl}$
		& \metric{1.0}{$1.50{\cdot}10^{-8}$} & \metric{1.0}{$1.18{\cdot}10^{-7}$} & \metric{1.0}{$1.39{\cdot}10^{-5}$} & \metric{1.0}{$1.48{\cdot}10^{-5}$} & - & - & \metric{1.0}{$2.71{\cdot}10^{-5}$} & \metric{1.0}{$3.04{\cdot}10^{-5}$} \\
		& $r_{vi}$
		& \metric{1.0}{$6.83{\cdot}10^{-4}$} & \metric{0.27}{$4.43{\cdot}10^{-1}$} & \metric{0.91}{$6.52{\cdot}10^{-3}$} & \metric{0.27}{$4.40{\cdot}10^{-1}$} & - & - & \metric{0.82}{$8.55{\cdot}10^{-3}$} & \metric{0.36}{$3.38{\cdot}10^{-1}$} \\
		\midrule
		\multirow{2}{*}{\textbf{Iron Man}}
		& $r_{J}$
		& \metric{1.0}{$4.43{\cdot}10^{-3}$} & \metric{1.0}{$1.91{\cdot}10^{-2}$} & \metric{0.42}{$9.59{\cdot}10^{-2}$} & \metric{0.87}{$1.68{\cdot}10^{-1}$} & \metric{0.5}{$7.16{\cdot}10^{-2}$} & \metric{0.88}{$1.59{\cdot}10^{-1}$} & \metric{1.0}{$4.24{\cdot}10^{-3}$} & \metric{1.0}{$9.38{\cdot}10^{-3}$} \\
		& $r_{kin}$
		& \metric{1.0}{$2.99{\cdot}10^{-3}$} & \metric{1.0}{$8.66{\cdot}10^{-3}$} & \metric{0.54}{$6.08{\cdot}10^{-1}$} & \metric{0.84}{$1.85{\cdot}10^{0}$} & \metric{0.54}{$6.10{\cdot}10^{-1}$} & \metric{0.85}{$1.82{\cdot}10^{0}$} & \metric{0.0}{$2.66{\cdot}10^{1}$} & \metric{0.06}{$3.97{\cdot}10^{1}$} \\
		\bottomrule
\end{tabular}
\end{table}

We further evaluate physical accuracy on full robotics models by measuring residuals along simulated trajectories. We compute configuration-level ($r_{J}$ and $r_{C}$), as well as velocity-level ($r_{kin}$ and $r_{v}$) residuals for joints and contacts. We further compute residuals of the contact NCP ($r_{primal}$, $r_{dual}$, $r_{cmpl}$, and the natural-map residual $r_{vi}$). Because most solvers do not expose internal joint forces, we must omit joint constraints from this second group of residuals. See Appendix~\ref{app:metrics} for a detailed description of these metrics. For BDX, Olaf, and DR~Legs, we use a contact-rich scenario in which each system is actuated with random torques, causing them to quickly fall to the ground. For Iron~Man, a fixed-base model, contacts are disabled and the model is simulated without actuation. In addition to comparing the results against the two GPU-based solvers (MJWarp and XPBD), we also compare the accuracy against the CPU implementation of MuJoCo, which can take advantage of 64-bit floating point arithmetic.

The results (\tabref{tab:perf-by-system-stats}) show that Kamino achieves the best or near-best residuals throughout, in particular on velocity-level metrics ($r_{kin}, r_{vi}$) evaluating the NCP solution accuracy. Joint constraint violations and contact penetration ($r_J$ and $r_C$) are also consistently low, although MJWarp's and MuJoCo's reduced-coordinate representation keeps joint constraints nearly exact on tree-structured systems like BDX. This supports the central claim that hard-constraint PADMM simulation improves the fidelity of the physics modeling for robotic systems with kinematic loops.

\subsection{Simulation Throughput}
\label{sec:results:performance}

To evaluate the throughput of Kamino, we consider three configurations afforded by the solver: a sparse formulation based on a matrix-free Conjugate Residual (CR) method applied to the Delassus operator, and two dense formulations based on block-Cholesky factorization (LLT) that differ in how the constraint Jacobian is represented (dense vs. sparse).

\begin{wrapfigure}[16]{R}{0.55\textwidth}
    \centering
    \vspace{-3.5ex} 
    \includegraphics[width=\linewidth]{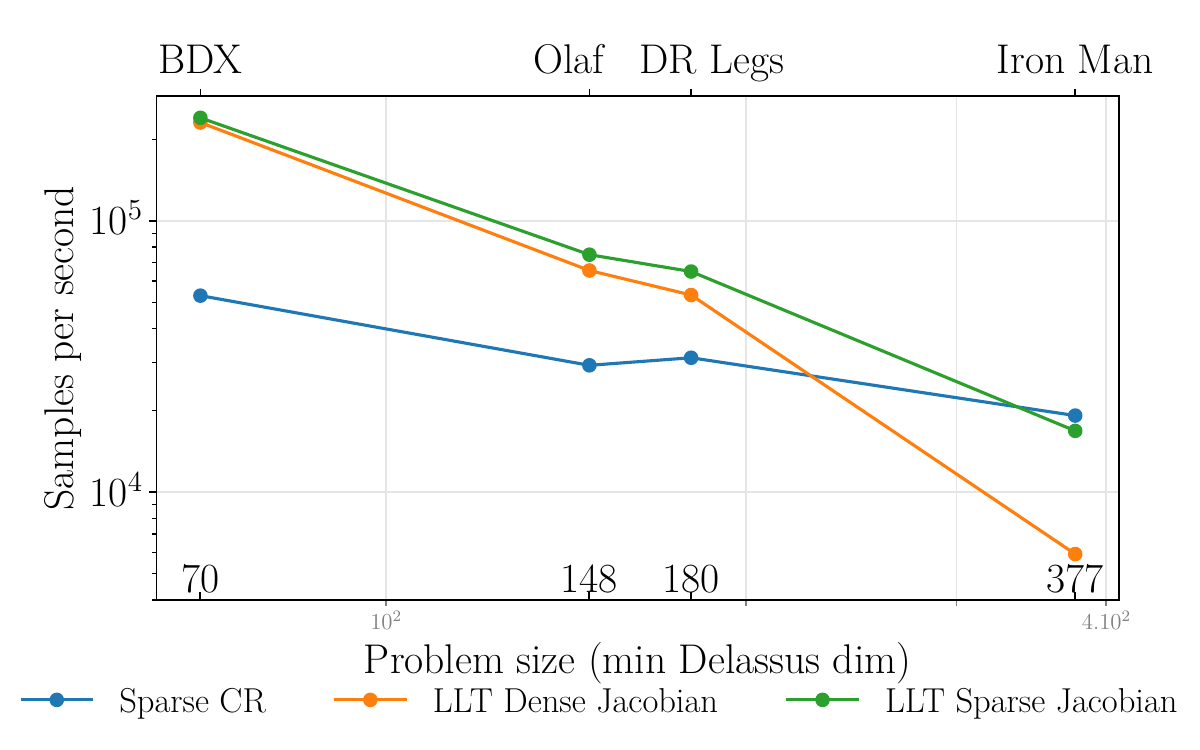}
    \caption{\textbf{Throughput.} Maximal simulator throughput vs. problem size across systems and solver configurations, measured on an RTX Pro 6000 GPU.}
    \label{fig:throughput}
\end{wrapfigure}

As shown in~\figref{fig:throughput}, the optimal solver depends on problem size. A dense solver (LLT with sparse or dense Jacobian) performs well on small systems, while the sparse CR solver scales well for larger models such as Iron~Man. In Appendix~\ref{app:throughput}, we additionally report results per model while varying the number of parallel worlds; as well as memory usage. The results confirm that the sparse Jacobian LLT consistently outperforms its dense variant.

We additionally compare throughput on BDX using Kamino (sparse LLT) against MJWarp and XPBD. Kamino achieves \num{2.4e5} samples per second, compared to \num{5.8e5} for XPBD and \num{6.1e6} for MJWarp.

\pagebreak
\subsection{Simulation of Complex Mechanisms}
\label{sec:results:simulation_examples}

We evaluate Kamino, MJWarp, and XPBD across three scenarios that highlight different aspects of rigid multi-body simulation with kinematic loops (see \secref{app:sim_models_overview} for model details and \secref{app:simulation_examples} for simulation configuration details).

\paragraph{DR Legs}
We simulate DR~Legs following a predefined gait using PD control. For this model, MJWarp requires an explicit articulation tree and loop-closure constraints. We evaluate two choices, rooted at the pelvis and at the left foot. Despite identical control inputs, only the pelvis-rooted configuration yields a stable walk, while the foot-rooted variant falls. Maximal-coordinate solvers are not affected by this artifact. Kamino produces a stable walking gait at a time step of $\Delta t = \SI{1.25}{\milli\second}$. In contrast, XPBD requires a reduced time step of $\Delta t = \SI{0.2}{\milli\second}$ and the animation to be played back at half speed to simulate a successful bipedal gait.

\paragraph{Iron Man}
This example, following a predefined animation using PD control, is challenging due to its many kinematic loops and large body mass ratios. As a result, MJWarp requires extensive tuning of loop-closure parameters to achieve a stable and accurate simulation, while XPBD needs $100$ iterations per simulation time step to avoid artificial compliance in the PD control. Both require a small simulation time step of $\Delta t = \SI{0.2}{\milli\second}$, whereas Kamino remains robust for a larger time step ($\Delta t = \SI{5}{\milli\second}$) with minimal tuning.

\paragraph{Hoberman Sphere}
Consisting of a very large number of bodies and kinematic loops, this model serves as a stress test for constraint resolution. We simulate the model spinning in zero gravity, leading to repeated expansion and contraction. While all solvers can simulate this scenario with a time step of $\Delta t = \SI{2}{\milli\second}$, MJWarp requires tuning of the loop-closure parameters for larger angular velocities, which in turn necessitates smaller time steps, whereas Kamino and XPBD can handle larger time steps more robustly.

Overall, we find that for models with kinematic loops, Kamino and XPBD require little parameter tuning. In contrast, MJWarp's tuning demands depend heavily on the scenario. Large velocities and loads, as well as large mass ratios, necessitate a more aggressive tuning of loop-closure constraints, which in turn requires smaller time steps for a robust simulation. Similarly, in the presence of contacts or PD control, XPBD requires more solver iterations and smaller time steps to achieve accurate results.

\subsection{Reinforcement Learning}
\label{sec:rl_results}

We demonstrate Kamino’s parallel simulation capabilities by training RL policies for BDX, Olaf, DR~Legs, and heterogeneous Ant variants, as shown in the supplementary video. We first validate performance on BDX and Olaf using the learning setups and reward formulations from their respective works~\cite{ruben2024design,muller2026olaf}. For BDX, which has a simple articulated topology, we observe policy performance similar to that of the original controllers trained in PhysX.
For Olaf, previous work~\cite{muller2026olaf} relied on a split-control architecture, where RL was applied only to the main robot structure (legs, torso, and head), while components with kinematic loops (shoulders, jaw) were controlled using joint-space PD control. In contrast, Kamino enables training a single end-to-end policy for the entire system. 
The RL policy trained for DR~Legs further demonstrates Kamino's ability to train policies for systems with complex kinematic loops. To assess reproducibility, we retrained the DR~Legs walking policy using $10$ independent random seeds. All seeds converge to the same gait, achieving an episode survival rate of $98.5 \pm 0.4\%$ with a reward variation of less than $2\%$. A detailed overview of the training setup for DR~Legs is provided in~\secref{app:dr_legs}.

To demonstrate Kamino's ability to simulate heterogeneous worlds, we consider a family of three variants of Ant: three-, four- and five-legged morphologies. Each world contains one of these three morphologies. A single policy is trained jointly across all variants to track target linear and angular velocity commands. Details are provided in~\secref{app:multi_ant}.

Finally, as depicted in the supplementary video, we deploy policies trained in Kamino on physical BDX and Olaf hardware, demonstrating successful sim-to-real transfer.


\section{Limitations}
\label{sec:limitations}

The current version of Kamino exhibits two main limitations.
First, on our current benchmarks, Kamino, which we built with high generality and accuracy as core design principles, generally achieves a lower throughput than MJWarp and XPBD. 
This is primarily due to our choice of directly solving the full NCP in maximal coordinates with PADMM, which requires repeated solves of a moderately large linear system.
Incremental engineering optimizations have, however, already brought substantial improvements to Kamino's performance without compromising accuracy; and future work on hybrid maximal-minimal-coordinate formulations could reduce the size of this linear system while retaining the benefits of our hard-constraint and hard-contact treatment.
Second, we have restricted our initial focus to linear 3D Coulomb friction, while torsional and rolling contact friction, as well as joint static friction, are not yet supported.

Finally, while not a limitation of Kamino itself, real-world validation in this work remains modest (sim-to-real transfer on BDX and Olaf). We plan to conduct more extensive evaluations in follow-up work, which could also inform the optimal choice of solver parameters and the refinement of heuristics such as warm-starting. We are nonetheless confident that our choice to place our simulator in the hands of the community will allow continuous improvements to the solver's performance, robustness, and scope of applicability.


\section{Conclusion}
\label{sec:conclusion}


We presented Kamino, a GPU-based physics solver for constrained rigid multi-body systems with complex kinematic topologies that combines a maximal-coordinate formulation with a PADMM solver. Its integration into Newton makes these capabilities available in massively parallelized simulation workflows, including heterogeneous worlds with structurally different robots.

Our experiments show that this formulation improves physical fidelity on elementary validation problems and applies to complex robotic mechanisms without the extensive example-specific tuning required by alternative formulations. Kamino retains this robustness while scaling to end-to-end RL training on thousands of parallel environments. We believe this combination is important for robot learning: it reduces the need to simplify mechanically rich robots into tree-structured approximations or extensively tune simulation parameters before they can be used in high-throughput training pipelines. Looking forward, the same PADMM perspective may provide similar accuracy and robustness benefits beyond the rigid-body systems studied here, including flexible bodies, transmissions, cables, and granular media.


\clearpage
\acknowledgments{The authors thank Sammy Christen, Josefine Klintberg, Espen Knoop and Luca Schwarzenbach at Disney Research for their contributions and feedback, Gilles Daviet, Eric Heiden, Miles Macklin, Gavriel State, Alain Denzler, Philipp Reist, Antoine Richard, Milad Rakhsha, Mohammad Mohajerani, Adam Moravanszky, and Spencer Huang at NVIDIA for their collaboration on the Newton framework and code reviews, and Yuval Tassa and Erik Frey at Google DeepMind for insightful discussions throughout the development of this work.}


\bibliography{main}  

\clearpage
\appendix
\counterwithin{figure}{section}
\counterwithin{table}{section}
\section{Appendix}
\label{app:appendix}

\subsection{Constrained Rigid Multi-Body Dynamics in Maximal Coordinates}
\label{app:dynamics}

To eliminate topological bias and natively support arbitrary kinematic configurations, Kamino adopts a maximal-coordinate formulation. Every rigid body $i \in \{1, \dots, n_b\}$ is represented with an independent configuration pose $\mathbf{q}_i = (\mathbf{r}_i, \boldsymbol{q}_i) \in \text{SE(3)}$, with orientation parameterized as a unit quaternion, and a six-dimensional spatial twist $\mathbf{u}_i = (\mathbf{v}_i, \boldsymbol{\omega}_i) \in \mathbb{R}^6$. Here $\mathbf{v}_i \in \mathbb{R}^3$ and $\boldsymbol{\omega}_i \in \mathbb{R}^3$ are the linear and angular velocity vectors of the respective body's CoM, expressed in world coordinates.

The generalized system mass matrix $\mathbf{M} \in \mathbb{R}^{6n_b \times 6n_b}$ is structured as a block-diagonal matrix:
\begin{equation}
\mathbf{M} = \text{diag}(\mathbf{M}_1, \mathbf{M}_2, \dots, \mathbf{M}_{n_b}),
\end{equation}
where each individual body mass matrix $\mathbf{M}_i \in \mathbb{R}^{6 \times 6}$ is defined using its scalar mass $m_i$ and its body-frame local inertia tensor $\mathbf{I}_i$ rotated into the global frame via the orientation rotation matrix $\mathbf{R}_i = \mathbf{R}(\boldsymbol{q}_i)$:
\begin{equation}
\mathbf{M}_i = \begin{bmatrix} m_i \mathbf{I}_3 & 0 \\ 0 & \mathbf{R}_i \mathbf{I}_i \mathbf{R}_i^\top \end{bmatrix}.
\end{equation}

Kinematic joint constraints, joint limits, and environmental contacts are all specified as algebraic constraints. Taking the first-order time derivative of the configuration-level constraints, $f(\mathbf{q}) = 0$, yields the velocity-level constraint equation $\mathbf{J}(\mathbf{q})\mathbf{u} = 0$, where $\mathbf{J} \in \mathbb{R}^{n_d \times 6n_b}$ collects joints $\mathbf{J}_J$, unilateral joint limits $\mathbf{J}_L$, and frictional contacts $\mathbf{J}_C$:
\begin{equation}
\mathbf{J} = \begin{bmatrix} \mathbf{J}_J^\top & \mathbf{J}_L^\top & \mathbf{J}_C^\top \end{bmatrix}^\top.
\end{equation}

\subsubsection{Signorini-Coulomb \& Bipotential Contact Model}
At an active contact $k$, the normal component obeys the Signorini condition: bodies may separate, and contact reactions may push, but a positive normal reaction can only occur when the contact is closed~\cite{moreau1988unilateral,glocker2001setvalued}:
\begin{equation}
v_{N,k}^+ \ge 0 \quad \perp \quad \lambda_{N,k} \ge 0.
\end{equation}
The tangential component is governed by Coulomb friction. For a given normal reaction $\lambda_{N,k}$, admissible tangential reactions lie in a disk of radius $\mu_k \lambda_{N,k}$. When the contact slips, the Maximum Dissipation Principle selects the reaction in that disk that removes the most tangential kinetic energy, i.e., the one opposing $\mathbf{v}_{T,k}^+$~\cite{moreau1977application}.

This coupling between the normal load and tangential dissipation makes the contact law non-associative: the feasible force cone is simple, but the velocity condition that produces maximally dissipative sliding is not the raw contact velocity. The de~Saxcé correction resolves this by adding a velocity-space offset to the normal component before testing cone complementarity~\cite{desaxce1991new,desaxce1998bipotential,acary2011formulation}. With $\mathbf{v}_k^+ = (\mathbf{v}_{T,k}^+, v_{N,k}^+)$, this gives
\begin{equation}
\hat{\mathbf{v}}_k :=
\mathbf{v}_k^+ + \boldsymbol{\Gamma}_{\mu_k}(\mathbf{v}_k^+)
\quad \in \quad \mathcal{K}_{\mu_k}^*
\quad \perp \quad
\boldsymbol{\lambda}_k \in \mathcal{K}_{\mu_k},
\qquad
\boldsymbol{\Gamma}_{\mu_k}(\mathbf{v}_k^+) =
\begin{bmatrix}
\mathbf{0}_{2}\\
\mu_k \|\mathbf{v}_{T,k}^+\|_2
\end{bmatrix}.
\end{equation}
Here, $\mathcal{K}_{\mu_k}$ is the 3D Coulomb friction cone and $\mathcal{K}_{\mu_k}^*$ is its dual cone. Intuitively, when a contact slides, the cone test receives an additional normal component proportional to the tangential slip speed. This makes the complementarity statement account for Moreau's Maximum Dissipation Principle, so a closed sliding contact neither separates nor penetrates as a byproduct of friction. Stacking these conditions across all joints, limits, and contacts gives the system-level inclusion
\begin{equation}
\mathbf{v}^+(\boldsymbol{\lambda}) + \boldsymbol{\Gamma}\big(\mathbf{v}^+(\boldsymbol{\lambda})\big) \in \mathcal{K}^* \quad \perp \quad \boldsymbol{\lambda} \in \mathcal{K},
\end{equation}
where $\mathcal{K} = \mathbb{R}^{n_{jd}} \times \mathbb{R}_{+}^{n_l} \times \mathcal{K}_{\boldsymbol{\mu}}$ bounds the admissible joint, limit, and contact reactions.

\subsubsection{Discrete-Time KKT Formulation and Dual Transcription}
Applying a semi-implicit Euler time-stepping scheme over a step $\Delta t$, or optionally a Moreau--Jean midpoint scheme evaluated at $\mathbf{q}^{k+1/2}$, converts the forward dynamics problem into the discrete KKT system~\cite{stewart1996implicit,jean1999nonsmooth}:
\begin{equation}
\label{eq:KKT_system}
\begin{bmatrix} \mathbf{M} & \mathbf{J}^\top \\ \mathbf{J} & -\mathbf{R} \end{bmatrix} \begin{bmatrix} \mathbf{u}^+ \\ -\boldsymbol{\lambda} \end{bmatrix} = \begin{bmatrix} \Delta t \, \mathbf{h}(\mathbf{q}, \mathbf{u}^-) + \mathbf{M} \mathbf{u}^- \\ -\mathbf{v}^* \end{bmatrix}.
\end{equation}
Here, $\mathbf{u}^-$ and $\mathbf{u}^+$ are the pre- and post-step generalized velocities, $\mathbf{h}$ collects external, gyroscopic, and applied control forces, and $\boldsymbol{\lambda} \in \mathbb{R}^{n_d}$ contains the constraint reaction impulses. The matrix $\mathbf{R}$ contains terms resulting from joint armature, viscous damping at joints, and implicit proportional-derivative (PD) actuator feedback, as detailed in Appendix~\ref{app:joint_dynamics}. The target velocity correction $\mathbf{v}^*$ collects Baumgarte stabilization, impact restitution, and reference actuator velocities.

The top row gives the post-step velocity as an affine function of the reactions:
\begin{equation}
\mathbf{u}^+(\boldsymbol{\lambda}) = \mathbf{u}^- + \Delta t \, \mathbf{M}^{-1}\mathbf{h} + \mathbf{M}^{-1}\mathbf{J}^\top \boldsymbol{\lambda}.
\end{equation}
Substituting this expression into the constraint row yields the Schur-complement form in reaction space. Combining it with the cone inclusion above gives \eqnref{eq:dual_problem} in \secref{sec:method}, the Nonlinear Second-Order Cone Program (NSOCP)
\begin{equation}
  \boldsymbol{\lambda} = \operatorname*{argmin}_{\mathbf{x} \in \mathcal{K}} \;
  \tfrac{1}{2}\, \mathbf{x}^\top \mathbf{D}\, \mathbf{x}
  + \mathbf{x}^\top \bigl(\mathbf{v}_f + \boldsymbol{\Gamma}(\mathbf{v}^+(\mathbf{x}))\bigr),
\end{equation}
with
\begin{align}
\mathbf{D} &= \mathbf{J} \mathbf{M}^{-1} \mathbf{J}^\top + \mathbf{R}, \\
\mathbf{v}_f &= \mathbf{J}(\mathbf{u}^- + \Delta t \, \mathbf{M}^{-1}\mathbf{h}) + \mathbf{v}^*, \\
\mathbf{v}^+(\mathbf{x}) &= \mathbf{D} \mathbf{x} + \mathbf{v}_f,
\end{align}
where $\mathbf{D}$ is the Delassus matrix, i.e., the apparent inverse inertia mapped into constraint space, and $\mathbf{v}_f$ is the unconstrained free constraint velocity. The operator $\boldsymbol{\Gamma}(\cdot)$ is zero for bilateral joints and unilateral limits and applies the de~Saxcé correction only to active frictional contacts. \citet{carpentier2024unified} provided a proof sketch of the equivalence between this NSOCP and the NCP describing rigid-body and hard-contact models. We refer readers to~\cite{tsounis2025solving} for the full derivation of the bipotential construction and the resulting dual dynamics problem.

\subsubsection{Joint-Space Dynamics}
\label{app:joint_dynamics}
For a single joint DoF $q$, we formulate a dynamics constraint derived from the single DoF equation,
\begin{equation}
 a \ddot{q} + b \dot{q} = -\Delta t^{-1}\lambda_q + \tau^\text{ff} + K_p (q^\text{ref} - q) + K_d(\dot{q}^\text{ref} - \dot{q}),
\end{equation}
with armature $a$, viscous damping $b$, PD gains $K_p, K_d$, feedforward torque $\tau^\text{ff}$ and reference trajectory $q^\text{ref}$, $\dot{q}^\text{ref}$. Each DoF can have a subset of these terms enabled. $\Delta t^{-1}\lambda_q$ represents the force or torque exchanged with the rest of the system.

Discretizing these equations with an implicit Euler scheme, we obtain
\begin{equation}
    a(\dot{q}^+ - \dot{q}^-) + \Delta t\, b\, \dot{q}^+ = -\lambda_q + \Delta t\left(\tau^\text{ff} + K_p (q^\text{ref} - q^- - \Delta t\, \dot{q}^+) + K_d(\dot{q}^\text{ref} - \dot{q}^+)\right).
\end{equation}
Factoring terms to isolate the unknown post-step joint velocity $\dot{q}^+$ gives
\begin{equation}
\left(a + \Delta t\, b + \Delta t\, K_d + (\Delta t)^2 K_p\right)\dot{q}^+ - \left(a\dot{q}^- + \Delta t\left(\tau^\text{ff} + K_p (q^\text{ref} - q^-) + K_d \dot{q}^\text{ref}\right)\right) = -\lambda_q,
\end{equation}
or, in compact form stacking all joint DoFs, with effective diagonal inertia $\mathbf{M}_q$ and bias $\dot{\mathbf{q}}^*$
\begin{equation}
\label{eq:joint_dynamics}
    \mathbf{M}_q (\dot{\mathbf{q}}^+ - \dot{\mathbf{q}}^*) = -\boldsymbol{\lambda}_q.
\end{equation}

Next, to relate the joint velocity to the body velocities, we use the kinematic relation
\begin{equation}
\label{eq:joint_kinematic_coupling}
\mathbf{J}_q \mathbf{u}^+ - \dot{\mathbf{q}}^+ = \mathbf{0},
\end{equation}
with Jacobian $\mathbf{J}_q$ mapping body to joint velocities. Combining \eqnref{eq:joint_dynamics} and \eqnref{eq:joint_kinematic_coupling} to eliminate $\dot{\mathbf{q}}^+$, we finally obtain the joint dynamics constraint
\begin{equation}
    \mathbf{J}_q \mathbf{u}^+ + \mathbf{M}_q^{-1}\boldsymbol{\lambda}_q - \dot{\mathbf{q}}^* = \mathbf{0},
\end{equation}
which has a form that aligns with the KKT system of \eqnref{eq:KKT_system} if one absorbs $\mathbf{J}_q, \boldsymbol{\lambda}_q, -\dot{\mathbf{q}}^*, \mathbf{M}_q^{-1}$, respectively, into $\mathbf{J}, \boldsymbol{\lambda}, \mathbf{v}^*, \mathbf{R}$.

\subsubsection{Physics Metrics}
\label{app:metrics}

We assess the modeling accuracy of a simulator by evaluating the residuals of the various constitutive equations along simulated trajectories of multiple robotic systems. Specifically, we consider the following metrics: 

\begin{itemize}[leftmargin=*]
\item the \textit{Constraint Violation Residuals}
\begin{equation}
r_{\text{J}}
:= 
\Vert\, \mathbf{f}_{J}(\mathbf{q}) \,\Vert_{\infty}
,\quad
r_{\text{C}}
:= 
\Vert\, \mathbf{f}_{C}(\mathbf{q}) \,\Vert_{\infty}
,
\end{equation}
where $\mathbf{f}_{J}$ and $\mathbf{f}_{C}$ are the total configuration-level constraint functions for joints and contacts, respectively. These metrics indicate the worst-case configuration-level joint violation and contact penetration occurring as a byproduct of velocity-level constraint violation integrated over time.
\item the \textit{Joint Kinematics Residual}
\begin{equation}
r_{kin}
:= 
\Vert\, \mathbf{J}_{J}(\mathbf{q}) \, \mathbf{u}^{+} \,\Vert_{\infty}
,
\end{equation}
where $\mathbf{J}_{J}$ is the joint constraint Jacobian evaluated at the pre-step initial body poses, and $\mathbf{u}^{+}$ is the post-event generalized velocity of the system. This metric serves as an indication of the per-step error contribution induced in the joint kinematic constraints by the forces and/or motion computed by the forward dynamics solver and the time integration scheme.
\item the \textit{Velocity-Level Contact Constraint Residual}
\begin{equation}
r_{v} := 
\max_{k \in \{1,\,\dots,\,n_{c}\}}{\,\,\max(0, -\mathbf{n}_{k}^{T}\,\mathbf{v}_{k}^{+})},
\end{equation}
representing the effective velocity-level constraint violation for contacts along respective contact normal directions, specifically in directions that would induce penetration. With $k$-subscripts denoting the contact index, $\mathbf{n}_{k}$ and $\mathbf{v}_{k}$ are the per-contact normal and relative contact velocity vectors, respectively, and $n_{c}$ is the total number of active contacts.
\item the \textit{NCP Primal Residual}
\begin{equation}
r_{\text{primal}} := \Vert \boldsymbol{\lambda} - \mathcal{P}^{\mathcal{K}}\left(\boldsymbol{\lambda}\right) \Vert_{\infty},
\end{equation}
representing the violation of set-valued constraint reactions, i.e. whether joint-limit and contact impulses lie within their respective cones.
\item the \textit{NCP Dual Residual}
\begin{equation}
r_{\text{dual}} := \Vert \hat{\mathbf{v}}^{+} - \mathcal{P}^{\mathcal{K^{*}}}\left(\hat{\mathbf{v}}^{+}\right) \Vert_{\infty}.
\end{equation}
This quantity indicates if non-zero velocities are present along directions that violate the constraint. For joints this simply amounts to the total constraint-space velocities, which should be zero at all times. For limits and contacts, it corresponds to non-zero velocities along the unilateral constraint dimension, e.g., the contact normal.
\item the \textit{NCP Complementarity Residual}
\begin{equation}
\boldsymbol{r}_{\text{cmpl}}
:= 
\left[
|\boldsymbol{\lambda}_{1}^{T} \, \hat{\mathbf{v}}_{1}^{+}|,
\dots,
|\boldsymbol{\lambda}_{k}^{T} \, \hat{\mathbf{v}}_{k}^{+}|,
\dots
\right],\quad k \in \{1, \dots, n_{c}\}
,\quad
r_{\text{cmpl}} := \Vert \boldsymbol{r}_{\text{cmpl}} \Vert_{\infty},
\end{equation}
which serves as a measure of adherence to the NCP. Non-zero values indicate a misalignment between constraint reactions and corresponding contact velocities along the tangential planes, i.e., the deviation from maximal dissipativity, as well as erroneous power distributions along contact normals.
\item  the \textit{NCP Natural-Map Residual} 
\begin{equation}
r_{vi} := \| \boldsymbol{\lambda} - \mathcal{P}^{\mathcal{K}}\left(\boldsymbol{\lambda} - \hat{\mathbf{v}}^{+}\right) \|_{\infty},
\end{equation}
 which indicates how close the computed contact reaction is to a solution of the NCP. Effectively, it combines the effects of the primal, dual, and complementarity residuals into a single value, providing an all-in-one measure of overall performance w.r.t. resolving the forward dynamics.\\
\end{itemize}

\subsection{PADMM Solver Details}
\label{app:solver}

PADMM solves the dual reaction-space problem by splitting it into two simple operations: a global quadratic update that accounts for the coupled dynamics through the Delassus matrix, and a projection that enforces membership in the feasible cone $\mathcal{K}$ defined in Appendix~\ref{app:dynamics}. The variables $\mathbf{x}$, $\mathbf{y}$, and $\mathbf{z}$ denote the unconstrained reaction update, the projected feasible reaction, and the ADMM dual variable, respectively; $\mathbf{z}$ also carries the velocity-like state used by the de~Saxcé estimate. This term is frozen at the previous iterate, so each PADMM iteration reduces to a linear solve followed by an analytic cone projection.

The resulting shifted Delassus system is $\mathbf{D}_{\eta,\rho} = \mathbf{D} + (\eta + \rho)\mathbf{I}$,
where $\eta$ is the proximal regularization and $\rho$ is the augmented-Lagrangian penalty. For fixed active constraints and solver parameters, $\mathbf{D}_{\eta,\rho}$ is constant throughout a time step. Kamino therefore performs the expensive factorization or sparse-operator setup once, then reuses it across all PADMM iterations while only changing the right-hand side. The per-iteration cascade is summarized in \algref{alg:padmm}.

\begin{algorithm}[tb]
\caption{Proximal-ADMM forward dynamics solver (one time step).}
\label{alg:padmm}
\begin{algorithmic}[1]
\REQUIRE Delassus $\mathbf{D}$, free velocity $\mathbf{v}_f$, feasible cone $\mathcal{K}$, parameters $\eta, \rho$
\STATE Build $\mathbf{D}_{\eta,\rho} = \mathbf{D} + (\eta + \rho)\mathbf{I}$
\STATE Initialize $\mathbf{x}^0, \mathbf{y}^0, \mathbf{z}^0$ from warm-start (\secref{app:warmstart})
\FOR{$i = 1, \ldots, N_{\max}$}
  \STATE $\mathbf{s}^i \leftarrow \boldsymbol{\Gamma}(\mathbf{z}^{i-1})$
  \hfill\COMMENT{de~Saxcé correction}
  \STATE $\mathbf{x}^i \leftarrow -\mathbf{D}_{\eta,\rho}^{-1}\bigl(\mathbf{v}_f + \mathbf{s}^i - \eta\, \mathbf{x}^{i-1} - \rho\, \mathbf{y}^{i-1} - \mathbf{z}^{i-1}\bigr)$
  \hfill\COMMENT{linear solve}
  \STATE $\mathbf{y}^i \leftarrow \mathcal{P}^{\mathcal{K}}\bigl(\mathbf{x}^i - \rho^{-1} \mathbf{z}^{i-1}\bigr)$
  \hfill\COMMENT{cone projection}
  \STATE $\mathbf{z}^i \leftarrow \mathbf{z}^{i-1} - \rho\,(\mathbf{x}^i - \mathbf{y}^i)$
  \hfill\COMMENT{dual update}
  \STATE Apply Nesterov acceleration to $(\mathbf{y}^i, \mathbf{z}^i)$
  \IF{\,\,$\max (r_\text{primal}^i, r_\text{dual}^i, r_\text{cmpl}^i) < \epsilon$\,\,}
    \STATE \textbf{break}
  \ENDIF
\ENDFOR
\RETURN $\boldsymbol{\lambda} \leftarrow \mathbf{y}^{N}$
\end{algorithmic}
\end{algorithm}

\subsubsection{Practical GPU Implementation}

Several implementation choices make this first-order method practical at simulation rates:

\paragraph{Nesterov Acceleration}
We accelerate PADMM convergence with Nesterov momentum applied to the consensus and dual variables~\cite{odonoghue2015adaptive}.
At each iteration, auxiliary variables $\hat{\mathbf{y}}^i$ and $\hat{\mathbf{z}}^i$ are computed by extrapolating from the current and previous iterates, with the momentum coefficient evolving as $a^{i+1} = (1 + \sqrt{1 + 4(a^i)^2})/2$.
Adapting the restart strategy of~\cite{odonoghue2015adaptive}, the acceleration is restarted (reset to $a = 1$) whenever the combined residual fails to decrease, preventing oscillation near the solution.

\paragraph{Warm-starting}
\label{app:warmstart}
Between time steps, the converged constraint reactions $\boldsymbol{\lambda}$ are cached and used to initialize the next solve.
For contacts, which may appear or disappear between steps, cached reactions are matched to the closest active contact by geometry pair and contact position.
%

\paragraph{Diagonal Preconditioning}
A diagonal Jacobi preconditioner $\mathbf{P} = \operatorname{diag}(\mathbf{D})^{-1/2}$ is computed once per time step.
The solver operates on the preconditioned system $\mathbf{P}\mathbf{D}\mathbf{P} + (\eta+\rho)\mathbf{I}$, which has improved spectral properties. This is especially relevant as Kamino uses single-precision floating point computations throughout (double-precision operations being significantly slower on the vast majority of recent GPUs).
%


\paragraph{Graph Capture} To eliminate kernel launch overhead, the entire PADMM solver loop (\algref{alg:padmm}) is captured as a CUDA graph using Warp's \texttt{wp.capture\_while}, which compiles the iterative loop into a single graph with conditional re-execution.
This is possible because all array shapes are fixed at graph capture time and the Delassus operator is constant within a time step. Additionally, a per-world mask allows early exit for worlds in which PADMM has already converged.
The captured graph is replayed each time step, amortizing the construction cost over the full simulation run.

\subsubsection{Linear System Solve}

The inner solve in \algref{alg:padmm} dominates runtime because it applies the globally coupled Delassus operator. Kamino supports two paths, matching the dense and sparse solver configurations evaluated in \secref{sec:results:performance}.

For small systems, Kamino explicitly assembles $\mathbf{D}_{\eta,\rho}$ and factorizes it once per time step using a block-Cholesky (block-LLT) decomposition. Each PADMM iteration then performs only the corresponding forward and backward substitutions with a new right-hand side.

For larger systems, assembling and factorizing a dense Delassus matrix becomes prohibitive in both memory and time. Kamino instead stores the constraint Jacobian in block-sparse-row (BSR) form. Since each constraint touches at most two bodies, each row has at most two nonzero $6$-coefficient blocks. The shifted operator is applied matrix-free as
\begin{equation}
\mathbf{D}_{\eta,\rho}\mathbf{v}
= \mathbf{J}\mathbf{M}^{-1}\mathbf{J}^\top\mathbf{v}
+ \mathbf{R}\mathbf{v}
+ (\eta+\rho)\mathbf{v}.
\end{equation}
Note that the product by $\mathbf{M}^{-1}$ is trivial in our maximal-coordinate setting as $\mathbf{M}$ is block-diagonal; in reduced coordinates this would require dedicated treatment \cite{sathya2026matrix}. The linear system is solved with a warm-started Conjugate Residual (CR) method~\cite{Macklin_2019}.

The matrix-free form is also useful if one wishes to update the penalty parameter $\rho$ adaptively, which can improve practical convergence \cite{boyd2011admm}. In dense form, updates to $\rho$ require an expensive matrix re-factorization, often offsetting the benefits of penalty adaptation.

Two GPU-specific choices reduce the sparse-solve overhead:
\begin{itemize}
    \item \emph{Pre-baked Jacobians}: the diagonal preconditioner and $\mathbf{M}^{-1}$ are folded into a copy of the constraint Jacobian at the start of each time step, avoiding separate scaling passes in every CR iteration.
    \item \emph{Low iteration budget}: each CR solve is initialized from the previous PADMM iteration's solution. Because successive right-hand sides are strongly correlated as ADMM converges, a small fixed budget is sufficient in practice; we use as few as $15$ CR iterations in the reported examples.
\end{itemize}

\subsection{Simulation Models Overview}
\label{app:sim_models_overview}

This section describes the simulation models used in our evaluation. Their structural properties are summarized in~\tabref{tab:systems}.

\begin{itemize}
    \item \textbf{DR~Legs}~\cite{gim2018design} is a bipedal robot with serial-parallel hybrid legs containing multiple four-bar linkages (\figref{fig:overview}). Each leg contains closed kinematic chains, resulting in $6$ independent loops. This model is used for physical validation, throughput evaluation, and reinforcement learning experiments (\secref{app:dr_legs}).
    \item \textbf{BDX}~\cite{ruben2024design} is a bipedal humanoid robot with a tree-structured kinematic topology and no closed loops. It serves as a baseline in the reinforcement learning experiments.
    \item \textbf{Olaf}~\cite{muller2026olaf} is a humanoid robot representing a well-known character. Its kinematic loops arise from mechanical couplings in the jaw and shoulder mechanisms.
    \item \textbf{Iron~Man} is a fixed-base Audio-Animatronics\textregistered\ figure with a densely connected closed-loop topology and strong mass heterogeneity across links. It contains $13$ kinematic loops and is used as a stress test for simulation stability under complex constraints.
    \item \textbf{Hoberman~Sphere} is a fully passive mechanism composed of repeated interconnected scissor-linkage loops. It is used as a constraint-handling stress test, containing $182$ kinematic loops.

    \item The \textbf{Manipulation} scenario uses a \emph{Franka Emika Panda} model, a 7-axis robot arm, with a \emph{Franka Hand}, a two-finger parallel gripper.
    \item \textbf{Ant~Variants} are three-, four-, and five-legged morphological variants of a standard quadruped robot. Each leg is actuated at the hip and knee joints. These variants are used in heterogeneous-world reinforcement learning experiments (\secref{app:multi_ant}), where a single policy is trained jointly across all morphologies.
\end{itemize}

\begin{table}[tb]
    \centering
    \caption{\textbf{Models Simulated with Kamino.} All counts refer to a single instance in maximal coordinates; loops denote independent closed kinematic chains.}
    \label{tab:systems}
    \footnotesize
    \setlength{\tabcolsep}{6pt}
    \begin{tabular}{l c c c c}
        \toprule
        \textbf{Model} & \textbf{Bodies} & \textbf{Actuated} & \textbf{Passive} & \textbf{Loops} \\
        \midrule
        DR~Legs          & $31$ & $12$ & $24$ & $6$  \\
        BDX              & $15$ & $14$ &  $0$ & $0$  \\
        Olaf             & $27$ & $20$ & $9$ & $3$  \\
        Iron~Man         & $67$ & $23$ & $57$ & $13$ \\
        Hoberman~Sphere  & $240$ & $0$ & $421$ & $182$ \\
        Ant-$3$            & $7$  & $6$  & $0$  & $0$  \\
        Ant-$4$            & $9$  & $8$  & $0$  & $0$  \\
        Ant-$5$            & $11$ & $10$ & $0$  & $0$  \\
        \bottomrule
    \end{tabular}
\end{table}

\subsection{Elementary Examples}
\label{app:elementary_examples}
\begin{figure}[tb]
    \centering
    \includegraphics[width=\linewidth]{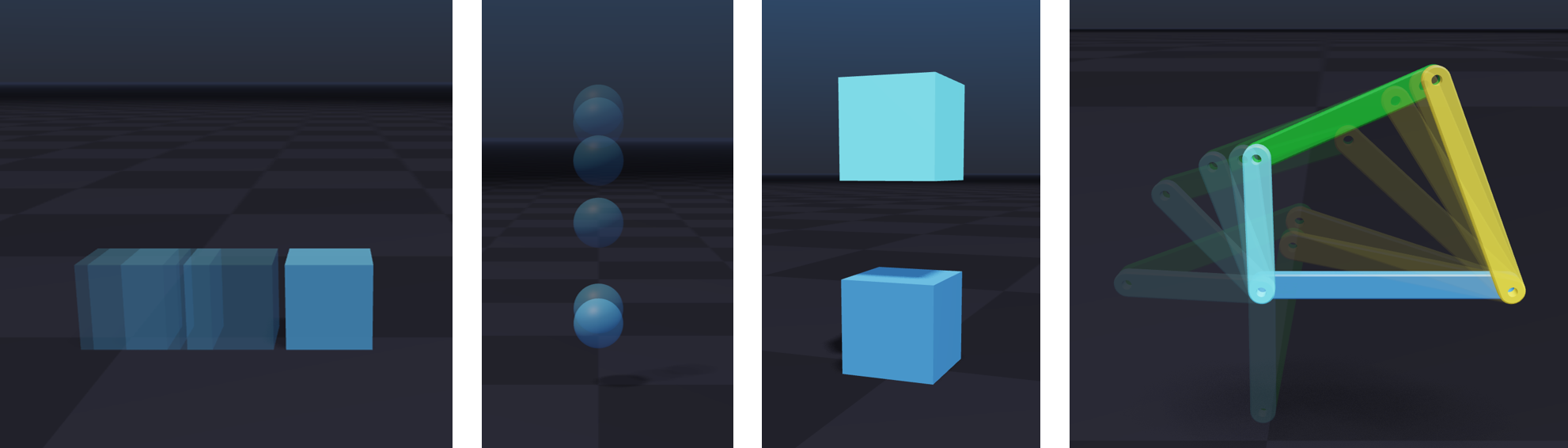}
    \caption{\textbf{Elementary physics validation examples.} Box on plane, bouncing ball, stacked boxes, and four-bar linkage.}
    \label{fig:validation_examples}
\end{figure}

We validate our mathematical modeling by assessing how well individual physical phenomena are captured compared to MJWarp and XPBD on four different examples shown in~\figref{fig:validation_examples}.

\paragraph{Box on Plane} In this first example, we consider a rigid box initially at rest on a flat plane. A horizontal force is applied and increased linearly from zero to $1.5$ times the theoretical stick-slip threshold, causing the box to slide. The force is then removed, and friction dissipates the kinetic energy until the box comes to rest again.

This serves as a validation example for hard contact modeling. In this simple case, the reaction forces predicted by the Signorini conditions, the Coulomb friction model, and the Maximum Dissipation Principle admit an analytical solution. A reference solution is generated by integrating Newton's second law twice using the analytical contact forces with a small time step (\SI{0.1}{\milli\second}). The trajectories predicted by the three solvers along the horizontal axis are compared against the reference.

\begin{figure}[tb]
\centering
\includegraphics[width=\linewidth]{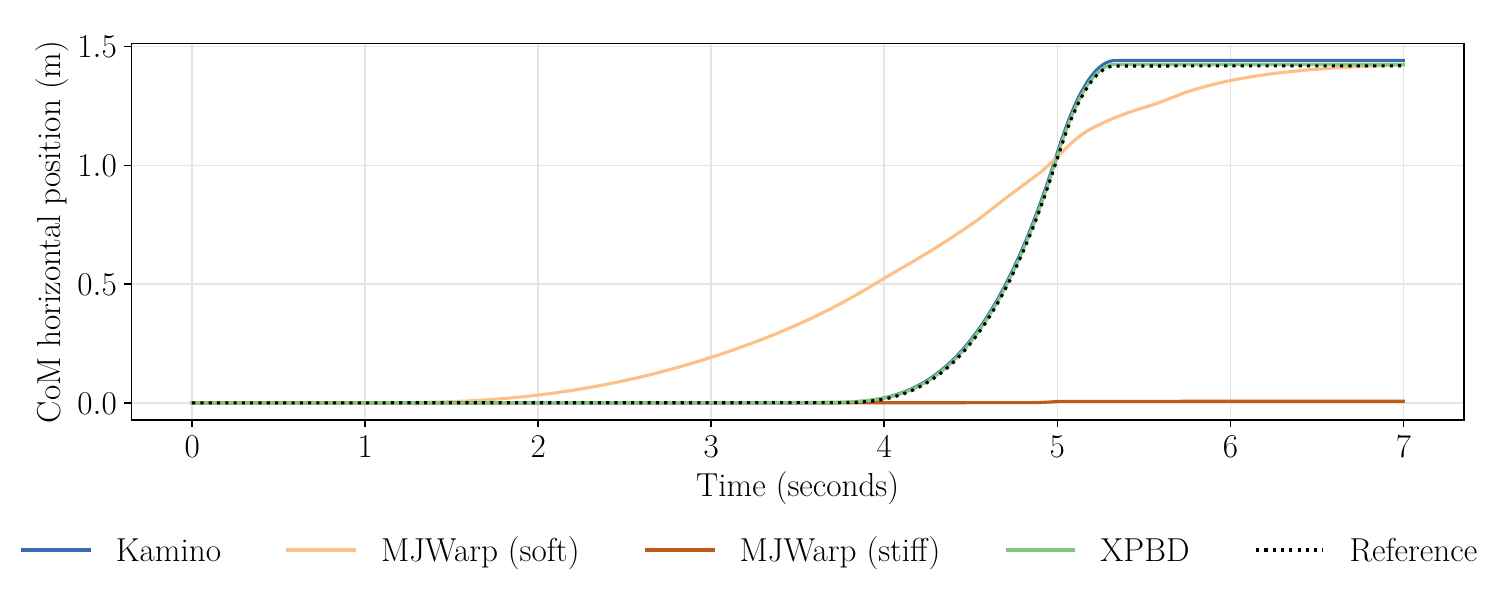}
\caption{\textbf{Box on Plane trajectory plots.} }
\label{fig:box_on_plane_plot}
\end{figure}

As seen in Fig.~\ref{fig:box_on_plane_plot}, both Kamino and XPBD produce a physically accurate trajectory (at the cost of increasing the number of XPBD iterations from the default $2$ to $20$). However, due to its formulation, XPBD does not produce exact sticking: the box starts sliding earlier than the reference solution, albeit very slowly, resulting in a smoother trajectory than expected.

MJWarp with default settings (``stiff" variant in the plot) exhibits excessive sticking. Its stick-slip behavior depends on solver parameters \texttt{ke}, \texttt{kd}, and \texttt{impratio}, which control constraint stiffness and damping in its softened contact formulation. While tuning these parameters can recover the correct total sliding distance (the ``soft" variant), the overall trajectory remains incorrect, as it requires overly compliant contacts that introduce noticeable ground penetration. Furthermore, the absence of the de~Saxcé correction in this solver leads to incorrect normal forces during sliding. As a result, the box does not slide smoothly, but wobbles as if the ground were uneven.

We also observe that, when the box is at rest, all solvers converge to a minimum-norm solution with equally distributed contact forces across the four box corners. In MJWarp and XPBD, this arises as a by-product of contact softening. In contrast, Kamino’s ADMM solver also produces these minimal-norm solutions, but without introducing additional softening or distributional bias.

\paragraph{Bouncing Ball} To evaluate the modeling of Newton’s law of restitutive impacts, we drop a ball from a fixed height onto a flat plane and simulate its vertical motion under gravity. The ball–ground restitution coefficient is set to $0.8$ to induce repeated impacts. We record the vertical position over time and compare it against the analytical solution, which consists of piecewise quadratic trajectories between consecutive impacts.

\begin{figure}[tb]
\centering
\includegraphics[width=\linewidth]{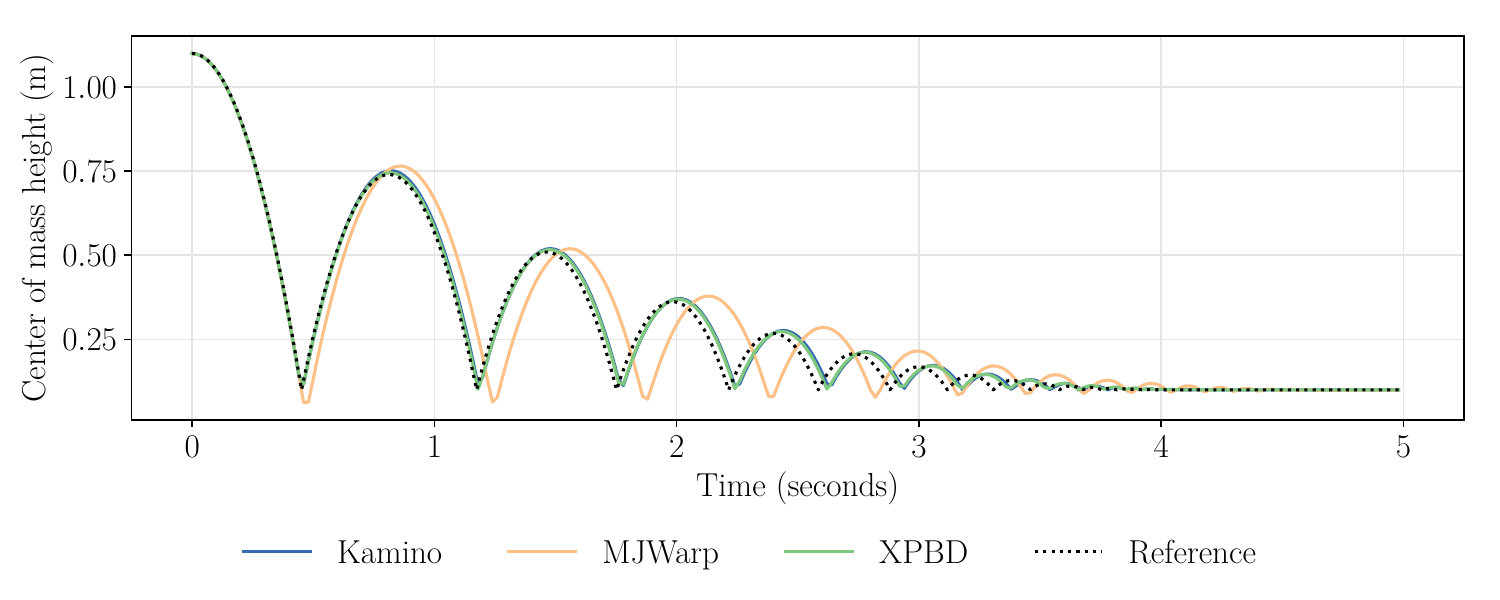}
\caption{\textbf{Bouncing Ball trajectories.} }
\label{fig:bouncing_ball_plot}
\end{figure}

As shown in~\figref{fig:bouncing_ball_plot}, Kamino and XPBD closely match the analytical trajectory. MJWarp does not natively support restitutive impacts. Instead, contact is handled through a compliant penalty formulation based on a linear spring–damper model, which can approximate this behavior when the contact is sufficiently under-damped. The effective restitution coefficient of this model can be derived from the spring–damper equations and used to compute suitable stiffness and damping parameters (\texttt{ke} and \texttt{kd}) for a desired impact response. This yields trajectories that are close to the analytical solution for the chosen restitution coefficient. However, this approximation introduces visible ground penetration due to under-damping and results in a coupling between physical and numerical parameters.

\paragraph{Stacked Boxes} In our next example, a rigid box is dropped vertically on top of another resting on a plane. Both boxes have identical dimensions, and we repeat the experiment with varying mass ratios. Note that, under such conditions, the analytical trajectory depends only on the initial height and the restitution coefficient and is independent of mass. This setup therefore evaluates the ability of solvers to handle varying mass ratios without introducing artifacts.

\begin{figure}[tb]
    \centering
    \includegraphics[width=\linewidth]{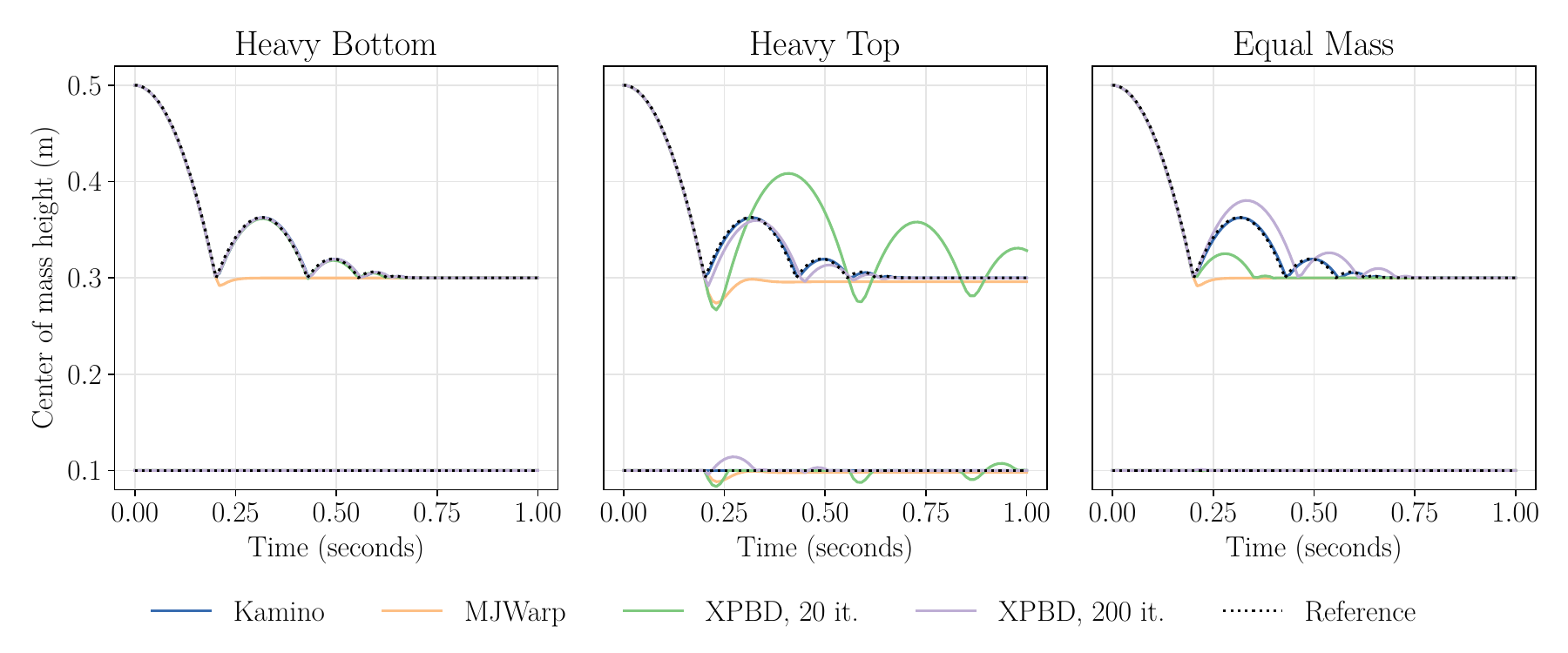}
    \caption{\textbf{Stacked Boxes.} Vertical position over time for a two-box system, where one box is dropped onto another resting on a plane. We evaluate three mass configurations: heavy bottom (\SI{100}{\kilogram}, \SI{1}{\kilogram}), heavy top (swapped masses), and equal mass (\SI{1}{\kilogram}).}
    \label{fig:stacked_boxes_plot}
\end{figure}

As can be seen in Fig.~{\ref{fig:stacked_boxes_plot}}, Kamino consistently predicts the correct solution regardless of the mass of the boxes. Due to constraint softening, MJWarp incurs some penetration before stabilizing, with larger penetration for a heavy top box. Note that we did not attempt to emulate restitution in this case; MJWarp's impacts are fully inelastic. Because XPBD relies on PGS-like iterations rather than on a global linear solve, its behavior depends on the mass distribution, with the heavy-bottom case being the most favorable. Increasing the number of iterations allows reducing the error with respect to the reference trajectory, but with a direct impact on time performance. Note that the error remains significant even with $200$ iterations per time step (the default setting being $2$ iterations). 

\paragraph{Four-Bar Linkage} We validate our handling of closed-loop kinematics on an elementary four-bar example --- the simplest non-trivial loop possible. All joints in this example are passive, and one of the bars is fixed. Released without initial velocity from an initial configuration, the linkage oscillates under gravity. We record position and velocity trajectories of all four bodies, and evaluate the total mechanical energy, as well as the position-level constraint residuals, along the trajectory.

\begin{figure}[tb]
\centering
\includegraphics[width=\linewidth]{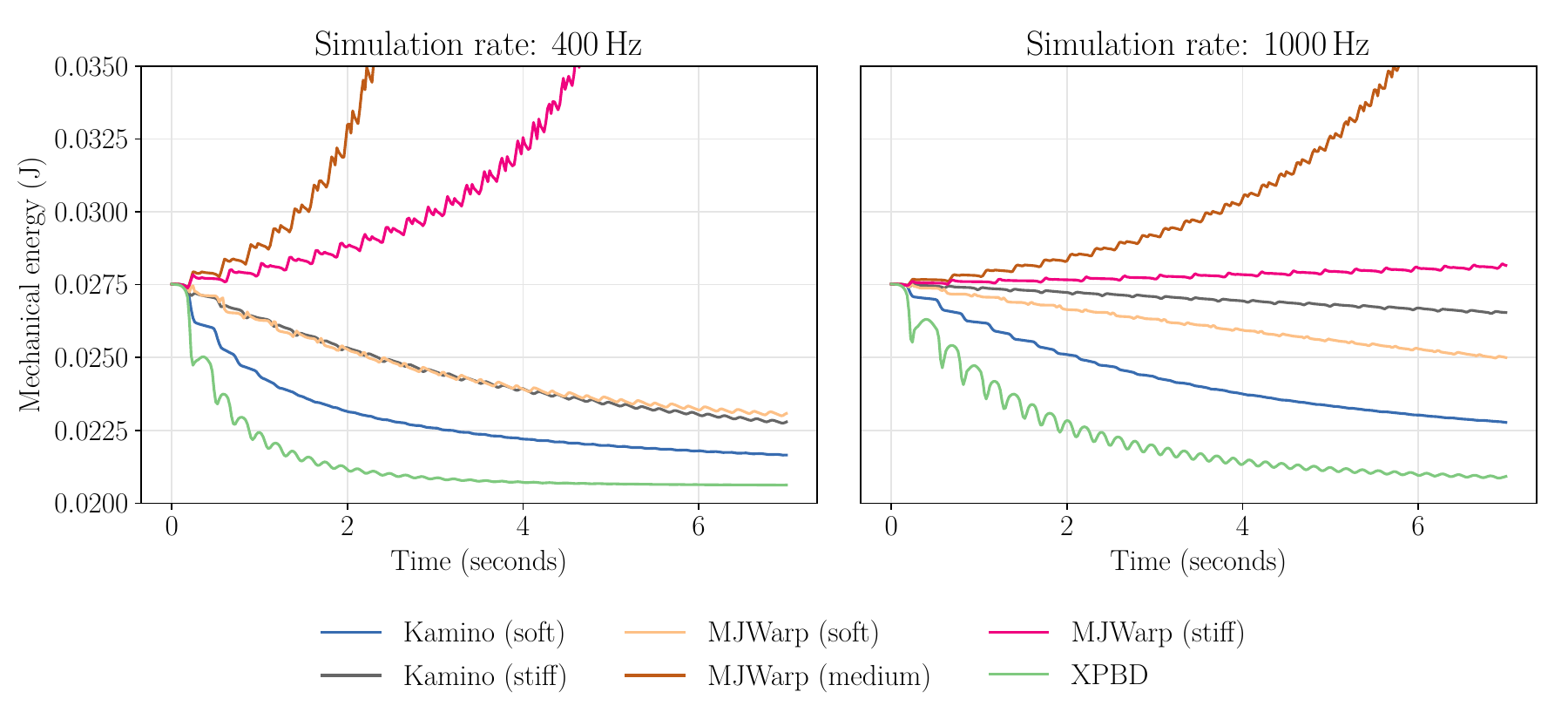}
\caption{\textbf{Four-Bar Linkage energy plots.}}
\label{fig:four_bar_plot}
\end{figure}

\begin{table}[tb]
\centering
\caption {\textbf{Four-Bar Linkage constraint residuals.} For each solver configuration, we report the maximal absolute position-level constraint residual over the trajectory (in meters for translation constraints, and in radians for rotation constraints).}
\label{tab:four_bar_constraints}
\renewcommand{\arraystretch}{1.15}
\setlength{\tabcolsep}{2pt}
\begin{tabular}{@{}l c c c c c c@{}}
\toprule
 \multirow{2}{*}[-0.5ex]{\textbf{Sim. rate}} & \multicolumn{2}{c}{\textbf{Kamino}} & \multicolumn{3}{c}{\textbf{MJWarp}} & \multirow{2}{*}[-0.5ex]{\textbf{XPBD}} \\
\cmidrule(lr){2-3}\cmidrule(lr){4-6}
 & Soft & Stiff & Soft & Medium & Stiff &  \\
\midrule
\SI{400}{\hertz} & $6.0{\times}10^{-3}$ & $1.8{\times}10^{-3}$ & $3.7{\times}10^{-3}$ & $9.4{\times}10^{-3}$ & $1.1{\times}10^{-2}$ & $1.4{\times}10^{-3}$ \\ 
\SI{1000}{\hertz} & $1.6{\times}10^{-3}$ & $3.6{\times}10^{-4}$ & $1.5{\times}10^{-3}$ & $1.8{\times}10^{-3}$ & $1.0{\times}10^{-4}$ & $2.2{\times}10^{-4}$ \\
\bottomrule
\end{tabular}
\end{table}

As we model all joints as ideal (no friction or damping), the mechanical energy is preserved in analytical solutions. Two parameters affect energy preservation and constraint satisfaction: the simulation time step, and the constraint stiffness (through the time constant $T$ in \texttt{eq\char`_solref} for MJWarp, the damping coefficient being set to $1$) or constraint stabilization weight $\alpha$ (for Kamino). We compare a \emph{soft} ($\alpha = 0.01$) and \emph{stiff} ($\alpha = 0.1$) version of Kamino, as well as a \emph{soft} ($T = 0.1$), \emph{medium} ($T=0.01$) and \emph{stiff} ($T=0.001$) version of MJWarp. For XPBD, we keep the default joint compliance of $0$ (hard constraints).

As can be observed in Tab.~\ref{tab:four_bar_constraints}, a small time step (\SI{1000}{\hertz}) and relatively stiff constraints are necessary to achieve sub-millimeter constraint satisfaction (the average link length being 
$\simeq$ \SI{6}{\centi\meter}). All solvers are dissipative for such a time step due to numerical integration, but to different extents, as shown by Fig.~\ref{fig:four_bar_plot}. XPBD and MJWarp achieve the tightest constraint satisfaction at \SI{400}{\hertz} and \SI{1000}{\hertz}, respectively, and XPBD is the most dissipative solver. Stiffer versions of Kamino and MJWarp dissipate energy slower, and even inject extraneous energy in the case of MJWarp, due to the non-linear spring dynamics at the constraints. Worse, this behavior is not monotonic, the version with medium stiffness being the one with the largest energy injection, presumably due to resonance. Only Kamino is reasonably dissipative, with a rate tied monotonically to the stabilization weight.

\subsection{Performance}
\label{app:throughput}
This section provides additional performance measurements of our physics solver, in particular regarding throughput and memory usage. We also detail how these vary w.r.t. solver configurations and hardware setups used for evaluation.

\paragraph{Throughput}
We measure our solver's throughput on challenging simulation sequences that test contact, joint limit and actuation paths. For BDX, Olaf and DR Legs, we apply random torques to all actuators, resampling and holding them for \SI{0.5}{\second} at a time, resulting in the systems falling quickly to the ground and hitting joint limits (5-13 contacts and 1-14 limits active per time step on average across models). For Iron Man, which is a fixed-base system (not expected to have active contacts) with many loops that could run into singularities if randomly actuated, we turn off all actuation and let the system passively collapse. To mitigate the effect of the particular sequence on solver convergence and thus indirectly on throughput, we disable any early stopping and always perform fixed PADMM and (when applicable) CR iteration counts (set to 100 and 15, respectively).

\begin{figure}[tb]
    \centering
    \includegraphics[width=\linewidth]{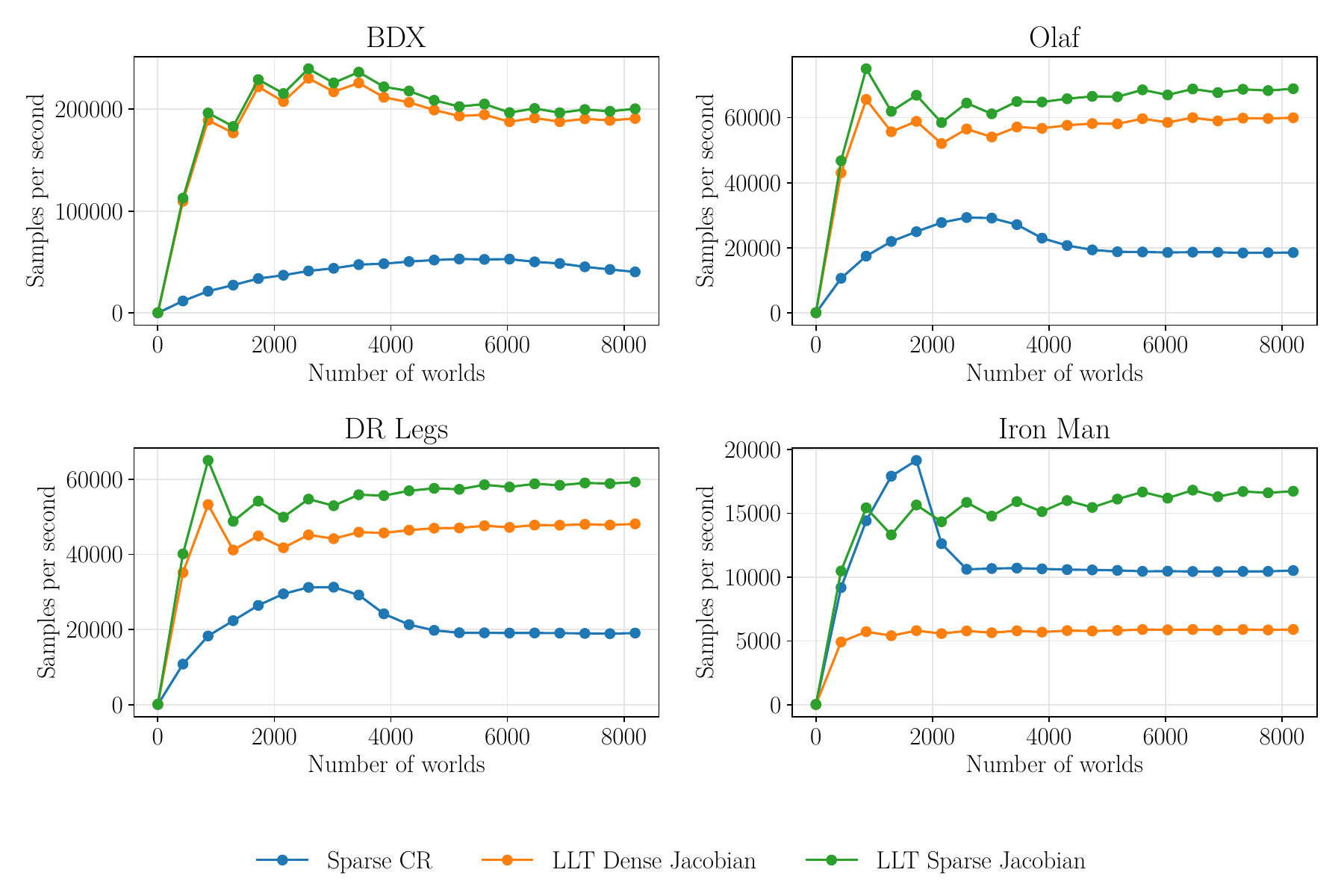}
    \caption{\textbf{Throughput scaling across solver configurations.} Simulator throughput (in samples per second) as a function of the number of parallel worlds for different problems and solver configurations. Measurements are performed on an RTX Pro 6000 Workstation GPU.}
    \label{fig:throughput_vs_worlds}
\end{figure}

In~\figref{fig:throughput_vs_worlds}, we report throughput as a function of the number of parallel worlds for each simulation model and solver configuration, illustrating scaling behavior under increased parallelization and GPU utilization. 

Since the size of the problem (e.g., the number of rows of the Delassus matrix) varies over time as contacts and constraints change, we use the minimal problem size, i.e., the case where no contact or joint-limit constraints are active, to report scaling results. This is a reasonable approximation in scenarios where the number of contacts remains low (e.g., walking). With RL applications in mind, we measure simulator performance as maximal throughput, defined as simulator steps per second times the number of parallel worlds.

We observe that throughput saturates once a few thousand worlds are simulated in parallel, as GPU compute resources become fully utilized. Beyond this point, the sparse CR solver path exhibits a noticeable decrease in performance, likely due to less regular memory access patterns. Note that the sawtooth patterns in the curves of the dense solvers are produced by the tail effect: the throughput drops when the number of scheduled threads just exceeds the Streaming Multiprocessor (SM) capacity, requiring an additional execution wave.

\begin{figure}[tb]
    \centering
    \includegraphics[width=0.7\linewidth]{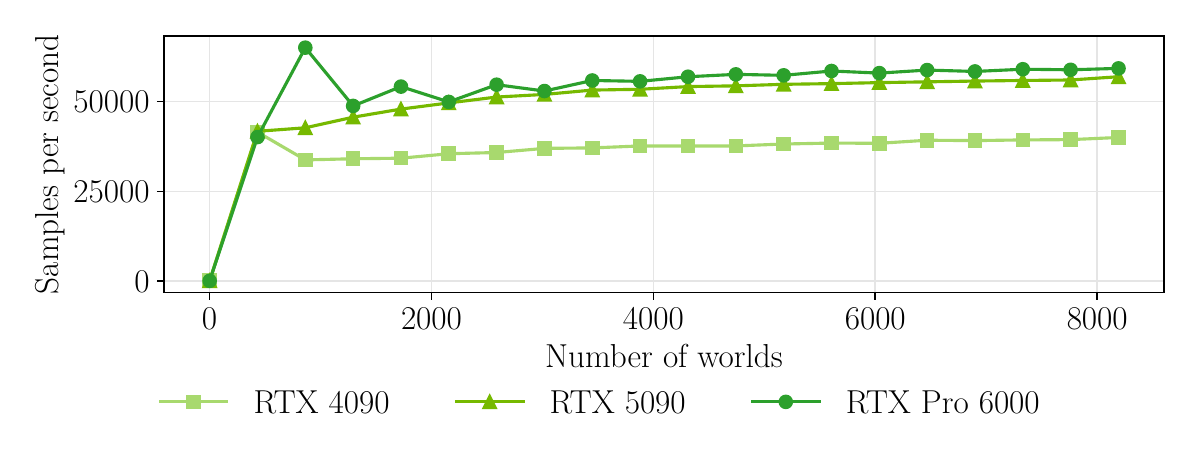}
    \caption{\textbf{Throughput across GPU hardware.} DR~Legs throughput scaling across different GPU models (RTX~4090, RTX~5090, RTX Pro~6000 Workstation Edition) using the best-performing solver configuration (LLT with sparse Jacobian).}
    \label{fig:hardware_comparison}
\end{figure}

In~\figref{fig:hardware_comparison}, we compare DR~Legs throughput across multiple GPUs (RTX~4090, RTX~5090, and RTX Pro~6000 Workstation Edition), providing a rough estimate of relative speedup across hardware configurations.

\paragraph{Memory Usage}
We also record GPU memory usage for the same benchmark trajectories under various solver configurations. Since all allocations are performed at the start of the simulation when creating the model and solver (no further allocations are performed when replaying the CUDA graph), we do not distinguish between average or peak memory usage.

\begin{figure}[tb]
    \centering
    \includegraphics[width=0.49\linewidth]{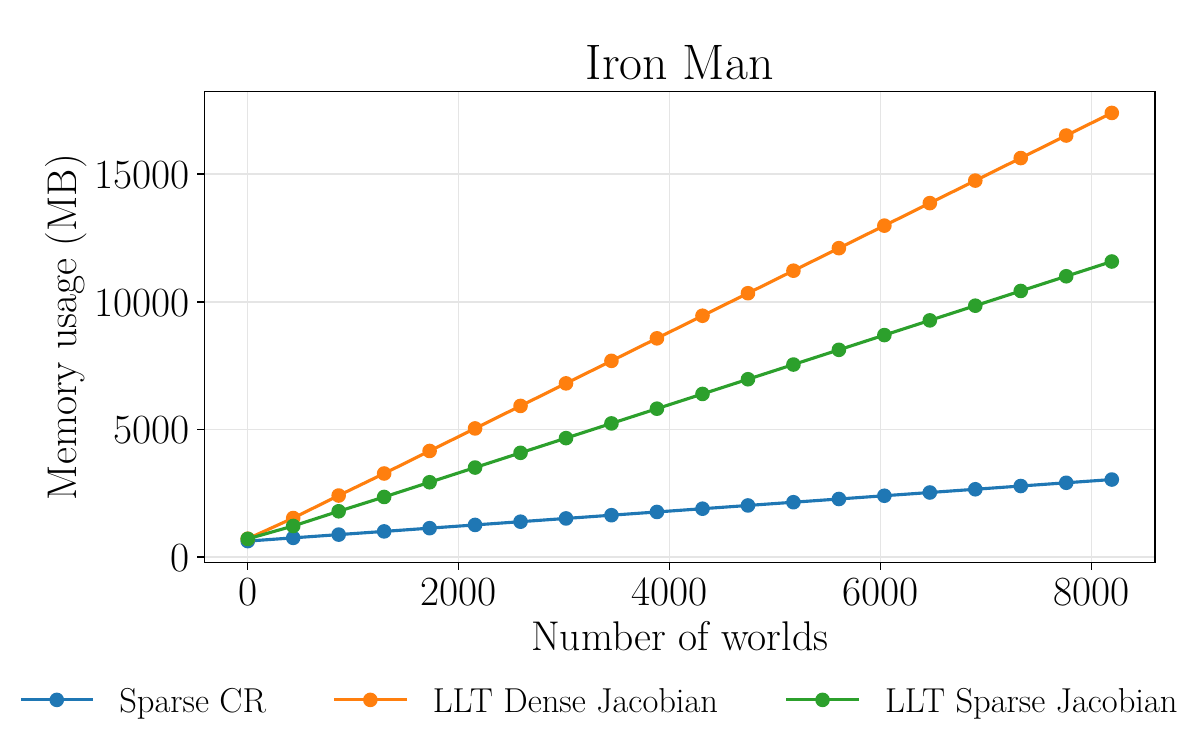}
    \includegraphics[width=0.49\linewidth]{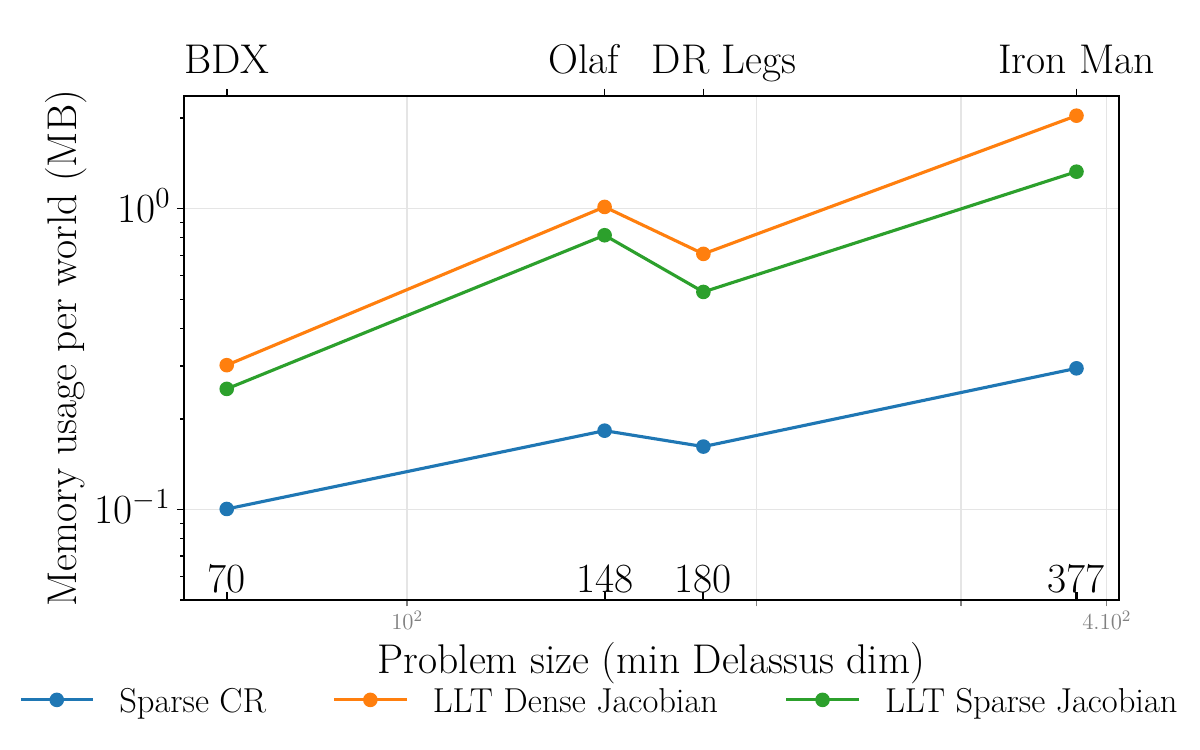}
    \caption{\textbf{Memory Usage.} Left: Total GPU memory usage for Iron~Man as a function of the number of parallel worlds. Right: GPU memory usage per world against problem size. Measurements are done on an RTX Pro~6000 GPU.}
    \label{fig:memory_usage}
\end{figure}

As can be seen in Fig.~\ref{fig:memory_usage} left, memory usage grows linearly with the number of worlds (for homogeneous worlds), with a constant term corresponding mainly to model geometry. The constant slope gives the per-world cost, which we plot against problem size in Fig.~\ref{fig:memory_usage} right. As expected, sparse solvers scale considerably better in terms of memory usage. Memory consumption is however not the main bottleneck for any of the systems we evaluated, even for dense solver variants.

\subsection{Complex Simulation Examples}
\label{app:simulation_examples}

We provide additional details on the simulation scenarios evaluated in \secref{sec:results:simulation_examples}.

\paragraph{DR Legs}
This simulation example uses DR Legs to follow a predefined gait animation, utilizing PD control through the solvers' implicit PD implementations. The animation is sampled at $\SI{100}{\hertz}$ and linearly interpolated for PD control setpoints. Articulation trees for MJWarp are generated through a breadth-first search starting at the root body, i.e. either the pelvis or the left foot, with the restriction that all actuated joints need to be part of the tree, i.e., loop-closure constraints need to represent non-actuated joints.

\paragraph{Iron Man}
The Iron Man example follows a predefined animation using PD control without contacts enabled. The animation is sampled at $\SI{100}{\hertz}$ and linearly interpolated for PD control setpoints. The model has $13$ kinematic loops that enable indirect actuation of its degrees of freedom, placing actuators where there is space, e.g., into the base for the actuation of the legs and into the torso for the actuation of the head. The masses of the moving rigid bodies range from $\SI{0.012}{\kilogram}$ to $\SI{49.20}{\kilogram}$, which can pose a challenge for simulations.

We tune the loop-closure constraint parameters for MJWarp to avoid excessive constraint violations, which necessitates a simulation time step of $\Delta t = \SI{0.2}{\milli\second}$ to ensure a stable simulation. For XPBD, the number of iterations is set to $100$, with a simulation time step of $\Delta t = \SI{0.2}{\milli\second}$, to avoid the artificial actuator compliance appearing at a low number of iterations or large time steps.

\paragraph{Hoberman Sphere}
We use the Hoberman sphere example to evaluate the simulation performance in the presence of many kinematic loops. The model is simulated in zero-gravity with an initial angular velocity of $\SI{0.5}{\radian\per\second}$. The initial velocity causes the sphere to expand, and with contacts disabled, the momentum of the expanding sphere results in a subsequent contraction after the sphere has passed the point of maximum expansion. The contraction will then once again turn into an expansion due to the angular velocity, creating a cycle. With the simulated initial conditions, maximum loop-closure constraint violations for MJWarp are only slightly larger than maximum constraint violations for Kamino and XPBD. However, with increased initial velocity or the presence of large external forces, tuning of loop-closure parameters becomes necessary to avoid excessively large constraint errors.

\subsection{Reinforcement Learning}
\label{app:rl}

\begin{figure}[tb]
    \centering
    \includegraphics[width=\linewidth]{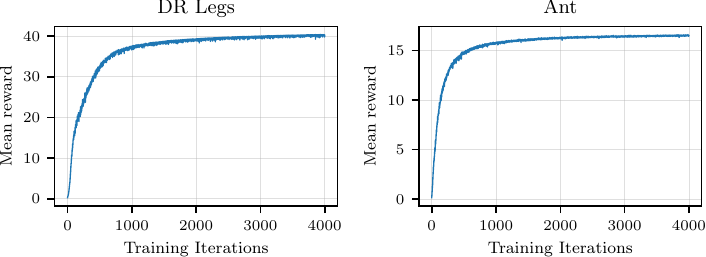}
    \caption{\textbf{RL training curves.} Mean episode reward for the DR~Legs task and heterogeneous Ant variants during training.}
    \label{fig:training_curves}
\end{figure}

We evaluate Kamino across several learning tasks. While for Olaf and BDX we follow the learning setup discussed in their respective papers, we establish new learning tasks for DR Legs and heterogeneous worlds (see learning curves in~\figref{fig:training_curves}). Below, we outline the specifics of these two learning tasks. In both cases, the policy operates at \SI{50}{\hertz} (simulator $\Delta t = \SI{4}{\milli\second}$, decimation $5$) and outputs joint-position targets for low-level Proportional-Derivative (PD) controllers across $4{,}096$ parallel worlds. The network architecture consists of a multi-layer perceptron with three hidden layers of $512$ units each and ELU activations, trained using Proximal Policy Optimization (PPO)~\cite{schulman2017proximal}. Training for $4{,}000$ iterations takes approximately \SI{3.5}{\hour} for DR Legs and \SI{2.8}{\hour} for the heterogeneous ant task on an RTX~5090 GPU.

\subsubsection{DR Legs}
\label{app:dr_legs}

We train a policy to track a base-frame SE(2) velocity command $(\hat{\mathbf{v}}_{xy,t}, \hat{\omega}_{z,t})$ on DR~Legs. The commands are sampled uniformly for each \SI{10}{\second} episode, with $\hat{v}_{x,t}, \hat{v}_{y,t}$ drawn from $\SI{\pm0.3}{\meter\per\second}$ and $\hat{\omega}_{z,t}$ from $\SI{\pm0.8}{\radian\per\second}$. To shape the contact patterns and swing-foot clearance, we employ a gait cycle with a period of $T = \SI{1}{\second}$. This choice closely approximates the characteristic period $2\pi\sqrt{L/g} \approx \SI{1}{\second}$ of a pendulum of length $L$, calculated at a standing height of $L = \SI{0.265}{\meter}$.

\paragraph{Action Space} 
Actions $\mathbf{a}_t \in [-1, 1]^{12}$ specify joint-position targets for the $12$ actuated joints. These targets are tracked using Kamino’s implicit PD actuator model, which integrates the PD dynamics directly into the constraint solver. The remaining $24$ passive joints that form the closed kinematic loops are left unactuated.


\paragraph{Observations} The robot's proprioceptive state is defined as
\begin{equation}
\mathbf{s}_t := (\mathbf{g}^\mathcal{R}_t, \boldsymbol{\omega}^\mathcal{R}_t, \hat{\mathbf{v}}_{xy,t}, \hat{\omega}_{z,t}, \boldsymbol{\phi}_t, \mathbf{q}_t, \mathbf{a}_{t-1}, \mathbf{a}_{t-2}),
\end{equation}
where $\mathbf{g}^\mathcal{R}_t$ denotes the gravity vector projected into the root frame and $\boldsymbol{\omega}^\mathcal{R}_t$ is the base angular velocity expressed in the root frame. The commanded base-frame velocities consist of the target planar linear velocity $\hat{\mathbf{v}}_{x,y}$ and target yaw rate $\hat{\omega}_z$. The gait phase is encoded as a sinusoidal embedding $\boldsymbol{\phi}_t = (\sin(2\pi t/T), \cos(2\pi t/T))$, where $T$ denotes the gait period. We further append the $12$ actuated-joint positions $\mathbf{q}_t$, as well as the actions from the two previous time steps, $\mathbf{a}_{t-1}$ and $\mathbf{a}_{t-2}$.

\paragraph{Rewards}
The reward terms (\tabref{tab:drlegs_walk_rewards}) are split into task tracking and regularization components. Task-tracking rewards encourage the robot to follow commanded base-frame velocities and to maintain a periodic walking pattern.
Using the gait period $T$, we define a linear phase of the gait as $\theta_t \in [0,1]$, where $\theta_t = (t / T) \bmod 1$. Based on $\theta_t$, we define a reference swing height for each foot $f \in \{L,R\}$ as $\hat{h}_f(\theta_t) = 0.05 \cdot \max(\sin(2\pi(\theta_t - \delta_f)), 0)$,
with phase offsets $\delta_L = 0.5$ and $\delta_R = 0$. 
The reference contact state is obtained by thresholding the reference swing height, $\hat{c}_f(\theta_t) = \mathds{1}[\hat{h}_f(\theta_t) < 10^{-4}]$, indicating whether the foot is in contact or in swing phase.
Additionally, the tracking rewards encourage an upright root orientation and maintain a nominal body height of \SI{0.265}{\meter}. Furthermore, both feet are aligned to a common yaw heading $\psi_{f,t}$, computed as the yaw angle of each foot's orientation quaternion.
Finally, a path deviation reward encourages the robot to remain close to the reference path position $\hat{\mathbf{p}}_{xy,t}$, obtained by integrating the commanded base-frame velocities, similar to \cite{ruben2024design}. 
Regularization terms penalize large root motion, high joint torques, and non-smooth actions, including action rate and action acceleration.

\begin{table}[htb]
\centering
\caption{\textbf{Reward terms for the DR~Legs walking policy.}
Hats $\hat{\cdot}$ denote reference quantities derived from the gait phase $\theta_t$. The indicator function $\mathds{1}[\cdot]$ evaluates to $1$ if the condition is satisfied and $0$ otherwise. The total reward is the weighted sum of all terms.}
\setlength{\tabcolsep}{6pt}
\begin{tabular}{l | c | l}
\toprule
\textbf{Name} & \textbf{Weight} & \textbf{Reward Term} \\
\midrule

\multicolumn{3}{c}{\textit{Tracking}} \\
\midrule

Linear velocity xy     & $5.0$        & $\exp\left(-\frac{1}{0.5^2}\,\lVert \mathbf{v}_{xy,t} - \hat{\mathbf{v}}_{xy,t} \rVert_2^2\right)$ \\
Angular velocity z     & $2.0$        & $\exp\left(-\frac{1}{0.5^2}\,(\omega_{z,t} - \hat{\omega}_{z,t})^2\right)$ \\
Path deviation         & $2.0$        & $\exp\left(-\frac{1}{0.1^2}\,\lVert \mathbf{p}_{xy,t} - \hat{\mathbf{p}}_{xy,t} \rVert_2^2\right)$ \\ 

Contact matching       & $10.0$        & $\sum_{f \in \{L,R\}} \mathds{1}[c_{f,t} = \hat{c}_f(\theta_t)]$ \\
Foot clearance         & $5.0$        & $\sum_{f \in \{L,R\}} \exp\left(-\frac{1}{0.02^2}\,(z_{f,t} - \hat{h}_f(\theta_t))^2\right)$ \\
Feet parallel          & $2.0$        & $\exp\left(-\frac{1}{0.2^2}\,(\psi_{L,t} - \psi_{R,t})^2\right)$ \\

Body height            & $1.0$       & $-(h_t - \SI{0.265}{\meter})^2$ \\
Upright orientation    & $5.0$       & $1-\lVert \mathbf{g}_{xy,t} \rVert_2^2$ \\
Survival               & $10.0$        & $1$ \\

\midrule
\multicolumn{3}{c}{\textit{Regularization}} \\
\midrule

Root velocity z        & $-2.0$       & $v_{z,t}^2$ \\
Root angular velocity xy & $-0.05$    & $\lVert \boldsymbol{\omega}_{xy,t} \rVert_2^2$ \\
Torque                 & $-0.001$    & $\lVert \boldsymbol{\tau}_t \rVert_2^2$ \\
Action rate            & $-0.01$        & $\lVert \mathbf{a}_t - \mathbf{a}_{t-1} \rVert_2^2$ \\
Action acceleration    & $-0.01$       & $\lVert \mathbf{a}_t - 2\mathbf{a}_{t-1} + \mathbf{a}_{t-2} \rVert_2^2$ \\

\bottomrule
\end{tabular}
\label{tab:drlegs_walk_rewards}
\end{table}

\paragraph{Episode Termination} An episode terminates after the \SI{10}{\second} time-out, when the base height drops below \SI{0.12}{\meter}, or when the base tilt with respect to the world-up axis exceeds \SI{1}{\radian}.


\subsubsection{Heterogeneous Ant Variants}
\label{app:multi_ant}

The base setup is derived from the MuJoCo Ant locomotion task. To demonstrate Kamino’s capability to simulate heterogeneous worlds, we train a single policy jointly across three-, four-, and five-legged Ant morphologies. Each simulated world contains one of these variants. The learning task is to track a base-frame SE(2) velocity command $(\hat{\mathbf{v}}_{xy,t}, \hat{\omega}_{z,t})$. The command is sampled uniformly with $\hat{v}_{x,t}, \hat{v}_{y,t}$ in $\SI{\pm0.3}{\meter\per\second}$ and $\hat{\omega}_{z,t}$ in \SI{\pm0.8}{\radian\per\second}.

\paragraph{Action Space}
Actions $\mathbf{a}_t \in \mathbb{R}^{10}$ specify target positions for joint-level Proportional-Derivative (PD) controllers. The action space is shared across all morphologies and sized according to the largest, five-legged variant. Each leg is actuated by a hip and knee joint, resulting in $10$ action dimensions in total. For the smaller three- and four-legged variants, the action dimensions corresponding to missing joints are ignored.

\paragraph{Observations}
The robot's proprioceptive state is defined as
\begin{equation}
    \mathbf{s}_t := (\mathbf{v}^\mathcal{R}_t, \boldsymbol{\omega}^\mathcal{R}_t, \hat{\mathbf{v}}_{xy,t}, \hat{\omega}_{z,t}, \mathbf{q}_t, \dot{\mathbf{q}}_t, \mathbf{a}_{t-1}, \mathbf{a}_{t-2}, \mathbf{c}_t),
\end{equation}
where $\mathbf{v}^\mathcal{R}_t$ and $\boldsymbol{\omega}^\mathcal{R}_t$ denote the base linear and angular velocities expressed in the root frame. The commanded linear velocity $\hat{\mathbf{v}}_{xy,t}$ is rotated into the root frame, while $\hat{\omega}_{z,t}$ specifies the target planar angular velocity. Joint positions $\mathbf{q}_t \in \mathbb{R}^{10}$ and joint velocities $\dot{\mathbf{q}}_t \in \mathbb{R}^{10}$ are defined using the dimensionality of the largest, five-legged morphology. For variants with fewer legs, entries corresponding to missing joints are set to zero. Finally, the policy is conditioned on the previous actions $\mathbf{a}_{t-1}, \mathbf{a}_{t-2}$ and a 3-dimensional one-hot encoding $\mathbf{c}_t$ indicating the active morphology.

\paragraph{Rewards}
The reward terms (\tabref{tab:multi_ant_rewards}) are divided into task-tracking and regularization components. The task-tracking terms reward accurate tracking of the commanded velocities. The regularization terms promote smooth and stable motion by penalizing hip yaw deviation, high joint velocities, and abrupt changes in the actions.

\begin{table}[H]
\begin{center}
    \caption{\textbf{Reward terms for the Ant variants policy.} Hats \(\hat{\cdot}\) denote reference commands. The total reward is the weighted sum of these terms.}
    \begin{tabular}{l | c | l}\toprule
    \textbf{Name}                  & \textbf{Weight} & \textbf{Reward Term} \\
    \midrule
    \multicolumn{3}{c}{\textit{Tracking}}  \\
    \midrule
    Linear velocity tracking       &  $10.0$  & $\exp\left(-\frac{1}{0.25^2} \cdot \lVert \mathbf{v}_{xy,t} - \hat{\mathbf{v}}_{xy,t} \rVert_2^2 \right)$ \\
    Angular velocity tracking      &  $10.0$  & $\exp\left(-\frac{1}{0.5^2} \cdot \lVert \omega_{z,t} - \hat{\omega}_{z,t} \rVert_2^2 \right)$ \\
    \midrule
    \multicolumn{3}{c}{\textit{Regularization}}  \\
    \midrule
    Hip angle penalty                &  $-1.0$  & $\lVert \mathbf{q}_\text{hip} \rVert_2^2$ \\
    Joint velocities               &  $-0.005$ & $\lVert \dot{\mathbf{q}} \rVert_2^2$ \\
    Action rate                    &  $-0.5$  & $\lVert \mathbf{a}_{t} - \mathbf{a}_{t-1} \rVert_2^2$ \\
    Action acceleration            &  $-0.5$  & $\lVert \mathbf{a}_{t} - 2\mathbf{a}_{t-1} + \mathbf{a}_{t-2} \rVert_2^2$ \\
    \bottomrule
    \end{tabular}
\label{tab:multi_ant_rewards}
\end{center}
\end{table}

\paragraph{Throughput}
We measure simulation throughput for $2{,}048$ environments for each individual morphology and for a heterogeneous mixture of all three variants. The results are summarized in \tabref{tab:multi_ant_throughput}. The heterogeneous configuration incurs some overhead compared to the four-legged variant, reflecting the additional cost of supporting multiple morphologies within a single simulation. Nevertheless, it remains faster than the five-legged variant, demonstrating that heterogeneous simulation can support multiple morphologies while retaining competitive simulation throughput.

\begin{table}[h]
    \centering
    \caption{\textbf{Simulation throughput for Ant variants.} Throughput is measured across $2{,}048$ parallel environments.}
    \label{tab:multi_ant_throughput}
    \begin{tabular}{lcccc}
    \toprule
    & 3-leg & 4-leg & 5-leg & Heterogeneous \\
    \midrule
    Samples per second & $286{,}510$ & $130{,}308$ & $53{,}360$ & $86{,}246$ \\
    \bottomrule
    \end{tabular}
\end{table}

\end{document}